\documentclass[conference]{IEEEtran}
\IEEEoverridecommandlockouts
\usepackage{cite}
\usepackage{amsmath,amssymb,amsfonts}
\usepackage{algorithmic}
\usepackage{graphicx}
\usepackage{csquotes}
\usepackage{textcomp}
\usepackage{hyperref}
\def\BibTeX{{\rm B\kern-.05em{\sc i\kern-.025em b}\kern-.08em
    T\kern-.1667em\lower.7ex\hbox{E}\kern-.125emX}}

\usepackage[most]{tcolorbox}

\usepackage{mdframed}
\usepackage{xcolor}

\definecolor{lightgray}{RGB}{220,220,220}

\newmdenv[linecolor=black,backgroundcolor=lightgray]{promptbox}

\usepackage{colortbl}
\usepackage{array,tabularx}
\definecolor{tablerow1}{RGB}{230,230,230}
\definecolor{tablerow2}{RGB}{245,245,245}

\usepackage{listings}
\usepackage{xcolor}

\definecolor{codegray}{gray}{0.9}
\definecolor{codepurple}{rgb}{0.58,0,0.82}
\definecolor{backcolour}{rgb}{0.95,0.95,0.92}

\lstdefinestyle{mystyle}{
    backgroundcolor=\color{backcolour},
    commentstyle=\color{codepurple},
    keywordstyle=\color{magenta},
    numberstyle=\tiny\color{codegray},
    stringstyle=\color{blue},
    basicstyle=\ttfamily\footnotesize,
    breakatwhitespace=false,
    breaklines=true,
    captionpos=b,
    keepspaces=true,
    numbers=left,
    numbersep=5pt,
    showspaces=false,
    showstringspaces=false,
    showtabs=false,
    tabsize=2
}

\begin{document}

\title{Semantic Compression With Large Language Models\\

}

\author{
\IEEEauthorblockN{Henry Gilbert, Michael Sandborn, Douglas C. Schmidt, Jesse Spencer-Smith, Jules White}
\IEEEauthorblockA{\textit{Dept. of Computer Science} \\
\textit{Vanderbilt University}\\
Nashville, TN, USA \\
\{henry.gilbert, michael.sandborn, douglas.schmidt, jesse.spencer-smith, jules.white\}@vanderbilt.edu}
}

\maketitle

\begin{abstract}
The rise of large language models (LLMs) is revolutionizing information retrieval, question answering, summarization, and code generation tasks. However, in addition to confidently presenting factually inaccurate information at times (known as ``hallucinations''), LLMs are also inherently limited by the number of input and output tokens that can be processed at once, making them potentially less effective on tasks that require processing a large set or continuous stream of information. A common approach to reducing the size of data is through lossless or lossy compression. Yet, in some cases it may not be strictly necessary to perfectly recover every detail from the original data, as long as a requisite level of semantic precision or intent is conveyed. 

This paper presents three contributions to research on LLMs.  First, we present the results from experiments exploring the viability of ``approximate compression'' using LLMs, focusing specifically on GPT-3.5 and GPT-4 via ChatGPT interfaces. Second, we investigate and quantify the capability of LLMs to compress text and code, as well as to recall and manipulate compressed representations of prompts. Third, we present two novel metrics---Exact Reconstructive Effectiveness (ERE) and Semantic Reconstruction Effectiveness (SRE)---that quantify the level of preserved intent between text compressed and decompressed by the LLMs we studied. Our initial results indicate that GPT-4 can effectively compress and reconstruct text while preserving the semantic essence of the original text, providing a path to leverage $\sim$5$\times$ more tokens than present limits allow. 
\end{abstract}


\section{Introduction}
\label{intro}

\textbf{Emerging trends and challenges.} Large language models (LLMs) have garnered significant attention with the recent release of OpenAI's ChatGPT \cite{chatgpt_site}, Google's Bard \cite{bard_site}, and others \cite{claude_site, amazon_site, ai21_site}. These LLMs facilitate (potentially incorrect) information retrieval, whether in the form of concept clarification, question answering, text editing, code generation, summarization, or task planning. LLMs typically provide users with an interactive chat interface to engage in back and forth conversation about a concept, task, or goal. These topics can be referenced over time within a conversation that preserves the discussion context over some time horizon (e.g., number of prompts or total tokens input in the session). 

A key to using LLMs effectively is the quality of the input \textit{prompt}, which is the text provided as input to the LLM. In conventional LLMs, the prompt is used in an opaque manner by the LLM to produce an output, called a \textit{response}, which is based on a non-linear function of the input prompt and the model's weights (the latter are opaque to users). Given some target task or output structural characteristics, the process of identifying high-quality prompts to feed an LLM is known as \textit{prompt engineering}~\cite{white2023prompt}. 

LLMs can be prompted with questions or directives, such as ``who was the first president of the United States?'' or ``provide a vegetarian dinner recipe that takes 20 minutes to prepare,'' and they often generate high-quality answers. A key limitation of conventional LLMs, however, is that they are trained on knowledge with a cut-off date.\footnote{ChatGPT's cut-off was September, 2021.} To reason or answer questions about newer information that an LLM was not trained on, the new information must be included in the prompt for the LLM. For example, a prompt could start with the text from a recent news article and then the LLM could be prompted with a question related to the information contained earlier in the prompt. The training cutoff increases the burden on the user to provide a sufficient context to the LLM about the nature of the desired output.

\textbf{Challenge: LLMs have a maximum input size, which restricts the amount of information that can be put into a prompt.} 
The maximum input size of an LLM is typically measured in tokens~\cite{openai_token_count, openai_token_limits} and corresponds to how much information can be input into the LLM at once. A token is a particular grouping of letters within a word. \textit{Tokenization} refers to how a given text is divided from letters and spaces into groups as a pre-processing step for an LLM, and several tokenization methods exist. This input token cap determines the maximum number of words or symbols that can be included in a prompt provided by a user. It also restricts the total amount of new information that can be provided to the LLM to perform a task, such as generating source code for software. 

GPT-3.5, which was the model underlying an earlier release of ChatGPT, has an input token limit of 4,096 tokens, or $\sim$3,000 words~\cite{openai_token_count}. The more recently released GPT-4 model has an increased input token limit of 32,768 tokens, or $\sim$24,000 words. It is likely, however, that LLMs will eventually operate on streams of data (e.g., daily customer activity or meeting transcripts) with much higher token counts. Moreover, while LLMs such as ChatGPT-4 can handle a conversation history, the same underlying token limit is applied. As such, the context the model is able to reason over is truncated by the internal token limit, regardless of prompt segmentation. 
% Guys, what sorts of use cases are you referring to here?  For example, the Chat GPT API
% "rate limits" programmatic requests, as per https://platform.openai.com/docs/guides/rate-limits/overview. Is this what you're referring to?

In many cases, the maximum input size is a significant limitation to the use of LLMs. For example, the source code for a large software application can often exceed the maximum input size to the LLM. As a result, the entire source code for the software cannot be provided to the LLM. It therefore may lack the full information needed to analyze the software thoroughly, generate new source code based on existing software components, and/or refactor the existing source code in a consistent and correct manner. 

Various approaches exist to overcome input size limitations, including using the LLM to summarize information that should be included in a prompt to shorten it. Other approaches range from (1) using semantic search to (2) only select contextually relevant information to include in a prompt to (3) structuring software into isolated components with clear interfaces that are easy to reason about using smaller segments of source code. Until the input size of LLMs becomes large enough to avoid becoming a concern in practice, however, determining how to use the limited space available in a prompt most efficiently remains an open research challenge.

 Motivated by these limitations, we examine the viability of prompting LLMs to produce compressed responses that still preserve rich semantic information, so that the original information can be recalled sufficiently and the original intent is preserved. Potential benefits of LLM compression capabilities include (1) large-scale, effective source code manipulation, (2) text retrieval, and (3) information distillation.

\iffalse
It is important to note that there are a number of concerns, known and unknown, about LLMs which center around bias, factual inconsistency, and interpretability of how an LLM maps a user input prompt to its generated response. The potentially harmful implications of these are many and range from information leakage to illusory logical conclusions. To our knowledge, there does not yet exist a reliable and robust framework or mechanism by which to evaluate or reason about the outputs produced by an LLM, making it a black-box model which simply produces the next predicted given the contextual tokens of a user-provided prompt. Often, a model presents seemingly plausible information which is only detectable by a highly attentive user or otherwise one with domain expertise in the prompt topic area. The research around evaluating LLMs along axes of bias mitigation, hallucination characteristics, and output interpretabilty, is still nascent.
\fi

\textbf{Research Question: Can LLMs perform their own compression on prompts to more efficiently embed information into prompts?} Based on the insight that LLMs can effectively summarize information to shorten it for inclusion in prompts, this paper analyzes whether LLMs can design and execute their own compression schemes to reduce the number of tokens consumed by a prompt.  In particular, we examine the ability of two different LLMs---GPT-3.5 and GPT-4---to achieve data compression and decompression on text and code. Compression is performed by asking the LLMs to compress provided prompts, whereas decompression is performed by entering the output from the LLM into a separated unrelated prompt to the LLM (e.g., outside of the prior conversation state) for use with subsequent reasoning. To limit human bias in our experiments, we directly request the LLM to produce a prompt that provides a directive to pass to other instances of the LLM to instruct them to perform compression on the subsequent data such that the compressed data can be effectively recalled later. 

This paper compares the compression rates, semantic similarity, and reconstruction and recall quality of GPT-3.5 and GPT-4 for compressed and uncompressed data between different conversations. We restrict the compression format to characters only (i.e., no emojis or Unicode characters). We evaluate the compression capabilities of these models against a standard compression algorithm, Zlib's Deflate algorithm~\cite{noauthor_2022-mu} using the compression ratio metric, edit distance metric, and our exact and semantic reconstruction effectiveness metrics. We use the cosine similarity between
embeddings\footnote{In this context, an embedding is a real-valued vector of reduced dimensionality derived from the input text data.} of compressed and decompressed text to quantify the quality of LLM compression. This work offers the following contributions to research on prompt engineering:
\begin{itemize}
    \item An initial exploration of LLM-based compression in GPT-3.5 and GPT-4, evaluated with two novel metrics: Semantic Reconstruction Effectiveness (SRE) and Exact Reconstruction Effectiveness (ERE)
    \item Experiments covering fictional and factual text obtained from the Internet
    \item Experiments covering code summarization and code generation tasks
    \item Exploration of LLM manipulation of compressed prompts for iterative construction of data
    % \item A discussion of ethical concerns and risks associated with model 
    % % Guys, should we provide a footnote defining what we mean by "model hypercommunication"?
    % hypercommunication, and
    \item Open source code of experiments and evaluation, which is available at 
    \url{https://github.com/henrygilbert22/phd\_chatgpt/tree/main/compression_analysis}
\end{itemize}

The remainder of this paper is organized as follows: Section~\ref{motivation}  provides background material that motivates our focus on LLM-based compression; Section~\ref{literary_text_performance} presents LLM compression experiments on excerpts from literary short story text; Section~\ref{results_prompt_eng} discusses prompt engineering approaches taken to facilitate compression; Section~\ref{results_code_integration} analyzes results from experiments using LLM compression for code generation; Section~\ref{additional} outlines additional studies and summarizes qualitative findings; Section~\ref{related} compares our approach with related work on evaluating LLMs and data compression;  and Section~\ref{conclusion} presents concluding remarks, lessons learned, and future research directions.

% %\input{sections/intro}
% \input{sections/related}
% %\input{sections/contributions}
% \input{sections/results_compression_performance}
% \input{sections/results_prompt_eng}
% \input{sections/results_code_integration}
% \input{sections/future}
% \input{sections/conclusion}

\section{Background on Compression vs. Embeddings}
\label{motivation}

% LLM-based compression can also apply to compressed code representations, enabling LLMs to generate or describe code based on a natural language description. Such capabilities may open new avenues in automated program repair applications, where LLMs can quickly identify and fix software defects subject to a dynamically updated test suite, thus yielding more dependable and secure software-reliant systems. While these applications of LLMs seem promising, we recognize the large gap between the interpretability of responses from existing LLMs and effectively leveraging functional approximate compression policies from these models. Our work therefore offers initial inslights into the viability of such compression policies.

% Another risk of LLM hypercommunication is the potential for these models to inadvertently leak sensitive information. As the information exchange rate of LLMs improves, there is an increased risk that private or confidential data may be compromised. This underscores the importance of developing robust security measures and data handling protocols to protect the privacy of users and organizations.

% \subsection{Embeddings vs. Compression}
% @Henry/Michael, please provide 1-2 paragraphs
% introducing the concept of embeddings and how
% they fit into the overall space before diving
% into the detailed discussion below. It isn't
% clear what the key points are in the material
% below relative to the overall goal of the paper.

Large language models (LLMs) are revolutionizing the field of natural language processing (NLP) by enabling more efficient and effective information storage and retrieval. A key research topic addressed by this paper is how well LLMs can perform their own prompt compression and manipulate compressed prompts, which enables the extraction and processing of information from shorter, condensed inputs while also maintaining semantic value and intent. A smaller prompt size may lead to faster response times and more efficient resource usage. 

We begin by highlighting the differences between embeddings and compression as they relate to the capabilites of LLMs. \textit{Embeddings}~\cite{google_embeddings} provide a one-way mapping for representing a high-dimensional, possibly sparse feature vector into a relatively low-dimensional space that still captures the semantics of the input. A common application of embeddings is to represent words of text \cite{mikolov2013efficient} or source code \cite{alon2018code2vec}. 

Embeddings are typically obtained by extracting the weights of a trained neural network layer that is close to the output layer, but before a classification layer 
(such as softmax \cite{10057747}) is applied. This approach captures the model's abstracted representation of the input that was learned implicitly during model training. Embeddings are inherently non-invertible\footnote{Non-invertible means that things like the original text cannot be recovered from the embedding of the text.}, owing to the information lost when reducing the dimensionality of the original input. In return, however, they offer the ability to quantify similarity of elements in the embedding space, typically with a vector-based metric, such as cosine similarity.  

In contrast, compression is concerned purely with minimizing the size of the data while also preserving the integrity of the information that is compressed. Depending on the algorithm used, compression may achieve perfect or near-perfect reconstruction of the compressed data into its original form. There are two main methods of compression: lossless and lossy \cite{adobe_compression}. Lossless compression removes only extraneous metadata, while lossy compression removes some of the original data. For example, PNG images use lossless compression, whereas JPEG images use lossy compression. Similarly, WAV is a lossless audio compression technique while MP3 provides lossy audio compresion. Lossy compression usually optimizes the information lost so that the decrease in quality is perceptually minimal to humans and is therefore tenable for most use cases.

Both embeddings and compression can be applied to a wide variety of media formats, ranging from classical text to images and three-dimensional modeling assets. To summarize, embeddings focus on transforming input into a different magnitude of dimensionality to facilitate comparison that is not easily achieved in the input space (e.g., semantic similarity of words or sentences). Conversely, compression focuses on minimizing the size of the input data so it can be reconstructed with original or near original fidelity. In other words, embeddings facilitate the quantitative comparison of data with unwieldy dimensionality, whereas compression minimizes the storage footprint of a piece of data while preserving as much of the original information as possible.

In the following sections, we report the results of our experiments exploring the following three key areas:
\begin{enumerate}
    \item Analysis of the compression rate and quality of fictional literary text 
    \item Prompt engineering tactics for eliciting compressed responses 
    \item Code generation capabilities with a compressed representation of a natural language description of a program. 
\end{enumerate}

We focus primarily on GPT-3.5 and GPT-4, which are the two models provided by OpenAI in a browser-based chat interface (i.e., ``ChatGPT'') with a ChatGPT Plus subscription. We leverage the API endpoints for GPT-3.5 where applicable, and otherwise produce prompts and responses for GPT-4 directly in the chat interface. The prompt text and results reside in our experiment notebooks available at \texttt{\small{https://github.com/henrygilbert22/phd\_chatgpt}}.
\section{Analyzing LLM Compression Performance on Literary Text }
\label{literary_text_performance}

This section provides initial results on experiments we conducted to evaluate potential applications and limitations of LLM compression. We use entropy, compression ratio, and edit distance to compare LLM compressed text to the baseline of Zlib's Deflate compression algorithm. Table~\ref{tab:text_ids} contains the names of the stories we studied, along with their associated identifiers used for brevity in corresponding figures.
% Texts studied table
\begin{table}
\caption{Text Excerpt Identifiers for the \\fictional short stories studied}    \label{tab:text_ids}
\centering
\begin{tabular}{|l|l|l|}

\hline

\textbf{Text ID} & \textbf{Text Name} & \textbf{Author} \\
\hline
a & A Good Man is Hard to Find & Flannery O'Connor \\
\hline
b & Break It Down & Lydia Davis \\
\hline
c & Cat Person & Kristen Roupenian \\
\hline
d & Cathedral & Raymond Carver \\
\hline
e & Flowers for Algernon & Daniel Keyes \\
\hline
f & Sticks & Karl Edward Wagner \\
\hline
g & Symbols and Signs & Vladimir Nabokov \\
\hline
h & The Bogey Beast & Annie Flora Steele \\
\hline
i & The Lottery & Shirley Jackson \\
\hline
j & The Veldt & Ray Bradbury \\
\hline
\end{tabular}
\end{table}

%\mbox{} \
\subsection{Experiment: Compression of Fictional Literary Text}
%\mbox{} \

We first compare the compression rate and reconstruction loss of GPT-4 with a standard compression method (Zlib Deflate compression algorithm) to evaluate how well GPT-4 performs with respect to compressing textual information effectively while simultaneously retaining semantic information. The evaluation set in this experiment is a collection of 10 excerpts from short story texts comprising a variety of genres and writing styles to assess model performance in compressing diverse text.

To initiate the compression process, we asked GPT-4 to generate a prompt that would facilitate text compression. By requesting that GPT-4 create its own prompt to facilitate compression and recall, we mitigated human bias and provided a standard prompt for our experiments. In response to this request, GPT-4 generated the following prompt for compression:

\begin{promptbox}
    \textit{Compress the following text into the smallest possible character representation. The resulting compressed text does not need to be human readable and only needs to be able to be reconstructed with a different GPT-4 model.}
\end{promptbox}

In a similar manner, we prompted GPT-4 to generate a prompt for text decompression. This prompt ensured that a new, independent GPT-4 model instance would decompress the compressed input text effectively. In response to this prompt, GPT-4 generated the following prompt for decompression tasks: 

\begin{promptbox}
    \textit{Please decompress the following compressed text into its original form, as it was provided by a user and compressed by another GPT-4 model.}
\end{promptbox}

Next we used these two prompts to instruct GPT-4 to compress each short story excerpt in the corpus. A separate instance of GPT-4 independent from the original model (i.e., in a separate chat conversation), was used to decompress the compressed text. By isolating chat conversations, we eliminated potential interference or leakage from the prompt to compress the original data. Prompt engineering tactics on compression quality are discussed further in Section~\ref{results_prompt_eng}.

To establish a baseline for comparison, Python's internal Zlib library \cite{python_zlib} using the Deflate and Inflate algorithms was used for both compression and decompression to assess existing lossless compression methods. We compressed each short story excerpt twice, first with minimal compression and second with maximal compression by passing the \texttt{level=0} and \texttt{level=9}, respectively. These results provide a basis to compare the effectiveness of GPT-4's compression capabilities.

%\mbox{} \
\subsection{Analysis: Entropy}
%\mbox{} \

Given the model's outputted compressed text, this section aims to model the derived information density and it's potential applications. To better understand the measure of randomness in the distribution of compressed characters, we calculate the frequency of each character in the compressed text and then the entropy of the distribution of compressed characters. Each compressed text was first converted to a byte stream representation before computing the distribution of its characters. The character distribution for a given text is given by:

\begin{equation}
P(x) = \frac{n_x}{N},
\label{eqn:char_prob}
\end{equation}

where $P(x)$ is the probability of character $x$, $n_x$ is the number of occurrences of character $x$, and $N$ is the total number of characters in the byte stream representation. The entropy was then computed for each byte character distribution using the entropy equation \cite{shannon_entropy}:

\begin{equation}
H(X) = -\sum_{x \in X} P(x) \log_2 P(x),
\label{eqn:entropy}
\end{equation}

where $H(X)$ is the entropy of the character distribution of a compressed text excerpt. 

\begin{table}
\centering
\caption{GPT-4 Compressed Text vs. Zlib Deflate By Entropy and Compression Ratio (CR)}
\label{tab:gpt4_comp_init}
\begin{tabular}{|c|c|c|}
\hline
% CR = compression ratio, ED = edit distance
\textbf{Method} & \textbf{Avg Entropy} & \textbf{Avg CR} \\
\hline
GPT-4 & 0.933 & 0.825 \\
\hline
Zlib Most & 0.837 & 0.469 \\
\hline
Zlib Least & 0.838 & 0.453 \\ 
\hline
\end{tabular}
\end{table}

\begin{figure}[tbph]
\vspace{-05pt}
\caption{Compression Entropy By Text Excerpt}
\centering
\includegraphics[width=0.45\textwidth]{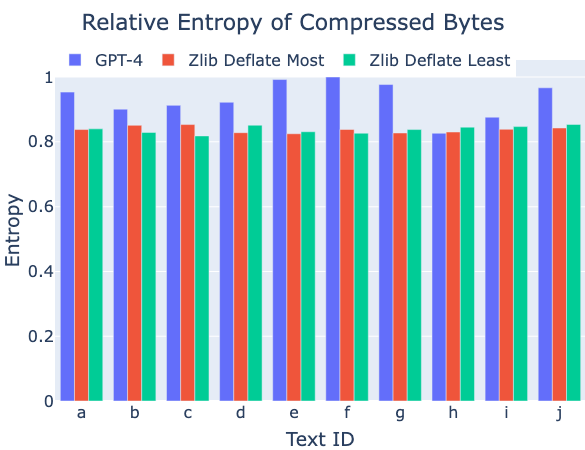}
\label{base_compression_entropy}
\end{figure}

Figure~\ref{base_compression_entropy} shows the entropy of each compressed text excerpt compared to the Zlib baselines, indicating that GPT-4 achieves the highest in resulting compression entropy for all texts tested.
 \mbox{} \

ChatGPT-4 large variation in produced compressed text may stem from it being trained on the largest corpus of data. As such, ChatGPT-4 implicitly reasons across a much larger dataset, giving it a larger potential distribution to derive outputs from. This would result in the derived distribution of compressed text to have a higher information density than comparable models. 

Table~\ref{tab:gpt4_comp_init} displays the averaged compression entropy and compression ratio for GPT-4 and the Zlib baselines.

\begin{figure}[htbp]
\caption{Averaged Compression Entropy By Text}
\centering
\includegraphics[width=0.5\textwidth]{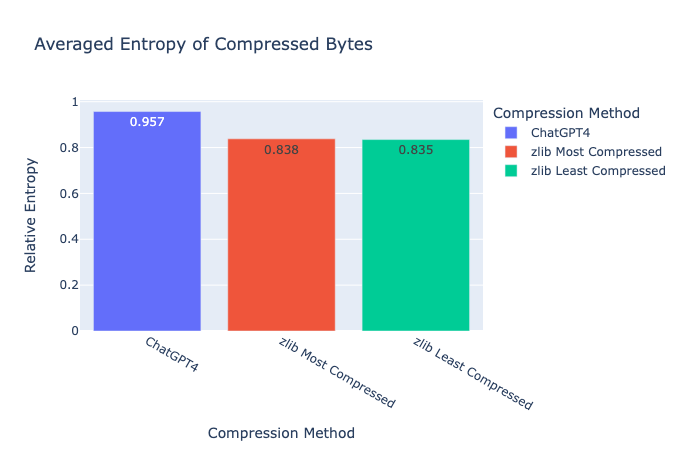}
\label{base_averaged_compression_entropy}
\end{figure} \mbox{} \

As anticipated, ChatGPT-4 consistently results in the highest entropy of its compressed text. In contrast, Zlib's most compression method generates slightly higher entropy than the least compression method. Given the small sample size of text excerpts and the small differences between these values, we cannot conclude that this trend holds with statistical significance. 

From our analysis we conclude that the entropy in the character distribution reveals GPT-4 consistently produces the highest entropy values for the text excerpts tested. This finding suggests that GPT-4 introduces more randomness in the compressed text compared to the Zlib baselines. Although higher entropy may be beneficial for certain applications, further research is needed to understand the implications of this characteristic.

%\mbox{} \
\subsection{Analysis: Compression Ratio}
%\mbox{} \

\begin{figure}[hptb]
\caption{Compression Ratio By Text}
\centering
\includegraphics[width=0.45\textwidth]{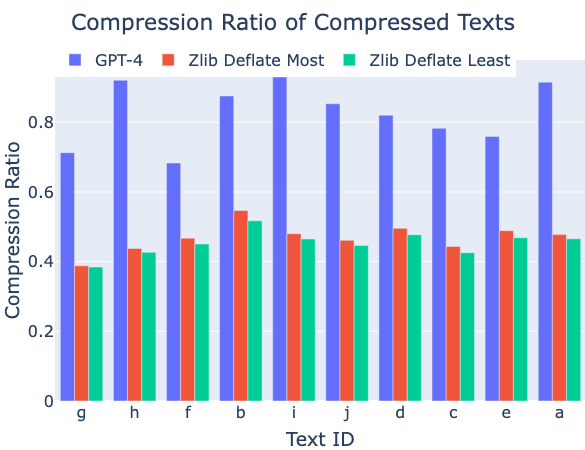}
\label{base_compression_ratio}
\end{figure}

To better understand the degree to which an input text was reduced in size, we compute the Compression Ratio (CR) between the original and compressed texts using the following equation:

\begin{equation}
CR = 1 - \frac{\text{\# compressed bytes}}{\text{\# original bytes}}
\end{equation}

where $CR$ is the compression ratio. For example, a compression ratio of 0.8 means that the original text size was reduced by 80\% in its compressed form, or is 20\% of the original text's size. Figure~\ref{base_compression_ratio} shows the relative compression ratio for each method across all texts.
 \mbox{} \
Clearly, GPT-4 provides higher compression ratios for the all of the text excerpts studied compared to the baseline methods. Zlib's most (\texttt{level=9}) aggressive compression method narrowly outperforms Zlib's least (\texttt{level=0}) aggressive method.

%This outcome is expected, as it is parameterized for maximal compression. The magnitude of performance change can likely be explained by the small size of the original text. 

% Figure~\ref{base_averaged_compression_ratio} displays the averaged compression ratio of each method across all texts.
% \begin{figure}[hptb]
% \caption{Averaged Compression Ratio By Text}
% \centering
% \includegraphics[width=0.5\textwidth]{figures/base_averaged_ratio.png}
% \label{base_averaged_compression_ratio}
% \end{figure} %\mbox{} \

As with Figure \ref{base_compression_ratio}, Zlib's maximal compression remains marginally better than Zlib's minimal compression. GPT-4 continues to outperform both baseline methods with a near 60\% increase in compression performance. While this result may initially appear remarkable, our subsequent analyses in Section~\ref{results_prompt_eng} reveal that GPT-4 achieved this high degree of compression rate by discarding key information in the original text.

% From this experiment, we conclude that ChatGPT-4's seemingly superior compression performance comes at the cost of losing information during decompression. This trade-off between compression rate and information loss must be carefully considered when determining the best compression method for a specific application. Further research is needed to optimize ChatGPT-4's performance while minimizing information loss, as well as to investigate other potential trade-offs between compression methods.

%\mbox{} \
\subsection{Analysis: Edit Distance}
%\mbox{} \

To better understand the exact closeness of the reconstructed text in relation to the originally compressed text, we use the Levenshtein edit distance metric \cite{lev_dist}:

\begin{equation}
\begin{split}
% D(i, j) = \begin{cases}
% 0 & \text{if } i = 0 \text{ and } j = 0, \
% i & \text{if } j = 0, \
% j & \text{if } i = 0, \
% \min \begin{cases}
% D(i-1, j) + 1 \
% D(i, j-1) + 1 \
% D(i-1, j-1) + (1 - \delta(s_i, t_j))
% \end{cases} & \text{otherwise},
% \end{cases}
D(i, j) = \begin{cases}
0 \quad \text{if } i = 0 \text{ and } j = 0, \\
i \quad \text{if } j = 0, \\
j \quad \text{if } i = 0, \\
\text{else} \\
\min \begin{cases}
D(i-1, j) + 1 \\
D(i, j-1) + 1 \\
D(i-1, j-1) + (1 - \delta(s_i, t_j))
\end{cases}
\end{cases}
\end{split}
\end{equation}

where $D(i, j)$ is the edit distance (number of characters that must be changed) between the original text of length $i$ and the reconstructed text of length $j$, $s_i$ and $t_j$ are the characters at position $i$ and $j$, respectively, and $\delta(x, y)$ is the Kronecker delta function which returns whether the characters at 2 possibly different indices are identical. All edit distances are normalized between 0 and 1, as the exact quantitative result is arbitrary in our case, and we only care about the relational analysis between methods. 

\begin{figure}[hptb]
\caption{Edit Distance By Text}
\centering
\includegraphics[width=0.45\textwidth]{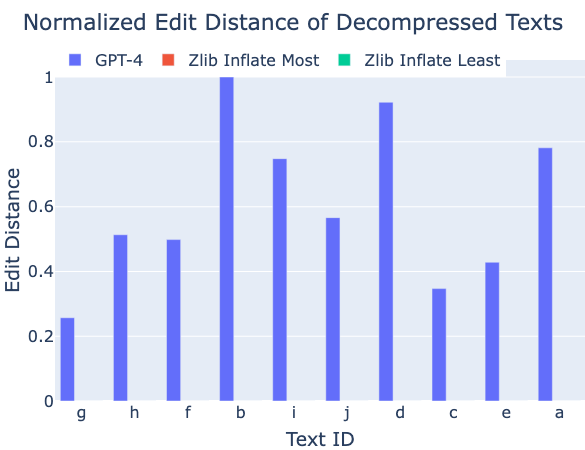}
\label{base_edit_distance}
\end{figure}

Figure~\ref{base_edit_distance} shows the edit distances for each compression method over all texts. 
 \mbox{} \
The Zlib baselines are lossless compression algorithms, so their edit distances are 0, as expected, and consequently not shown on the graph. From this figure we observe that GPT-4 compression is rather lossy. 

Figure~\ref{base_edit_distance} also showcases the variance of exact reconstruction performance. In particular, GPT-4's performance varies greatly on different text excerpts. This edit distance variance in GPT-4's compression quality raises questions for future studies, such as what character distributions and captured features of input text influence a higher or lower edit distance in the compressed text. The reasons behind this variance could be due to inherent characteristics of the input text, such as language structure, semantic complexity, or specific dialect and choice of language. This variance requires further investigation into the factors influencing GPT-4's performance on text compression and decompression tasks.

The largest performance difference is $10\times$ the smallest, correlating loosely with the compression ratio and compressed text distribution entropy. The average of these edit distances is 0.369. Note that the concrete edit distance is arbitrary without a baseline to compare against. 

% \begin{figure}[h!]
% \caption{Edit Distance By Text}
% \centering
% \includegraphics[width=0.5\textwidth]{figures/averaged_base_edit.png}
% \label{base_averaged_edit_distance}
% \end{figure} \mbox{} \

% Figure \ref{base_averaged_edit_distance} shows the averaged values of edit distances across all texts by compression method. Naturally, the zlib methods remain 0 as they are lossless. The averaged normalized edit distance value of ChatGPT-4 gives minimal additional value without the relational context.

% Future research should explore the relationship between the input text and ChatGPT-4's compression performance, as well as ways to optimize the model's ability to maintain lossless compression without sacrificing compression ratios. Additionally, it would be valuable to investigate how other large language models compare to ChatGPT-4 in terms of compression performance and edit distance variability.

%\mbox{} \
\subsection{Analysis: Semantic Retention Quantified by Cosine Similarity}
%\mbox{} \

%this is not what large language models (LLMs) are made for, as we are not utilizing their core ability to abstractly reason about relations. Thus, rather than measure the exact character difference,

Based on previous results, GPT-4 cannot currently be used as a reliable compression technique since it does not rival existing lossless methods based on the edit distance metric. This finding indicates that information is lost between compression and decompression when input text is passed to the LLM to reduce its size. However, we nevertheless want to explore the ability of an LLM to capture the \textit{underlying semantic intent} of the original text in an approximately reconstructable manner. 

We are not concerned whether the decompressed text exactly matches the original, as long as it retains the essence of what is originally intended to be communicated. For example, if the original message is: ``Please send me an email on Monday'', and the reconstructed message is: ``On Monday, send me an email, please'', then the resulting semantic similarity score should be relatively high as the underlying meaning of ``send an email Monday'' is represented in both messages, even though the character occurrence and alignment do not closely match.

Quantitatively evaluating whether two texts have the same semantic meaning is non-trivial and slightly arbitrary since traditional text comparison methods do not apply. To quantify similarity of compressed and decompressed texts, we use OpenAI's Embeddings API \cite{openai_emb}, and then apply the cosine similarity vector metric to pairs of embedded texts. Section~\ref{intro} explained how embeddings are vectors that represent (1) the semantic information in the text learned by an LLM during its training and (2) a high-dimensional, possibly sparse vector in a low-dimensional representation. A text embedding captures semantic meaning in the embedding space, enabling similarity comparisons using vector-based metrics. The commonly used metric for embedding comparison is cosine similarity \cite{cos_sim}, which is calculated as follows:

% We can now directly measure our divergence from the original message's semantical meaning using the cosine similarity between the original text's embedding vector and the decompressed text's embedding vector. 

\begin{equation}
\label{eqn:cos_sim}
\text{Cosine Similarity}(\mathbf{A}, \mathbf{B}) = \cos (\theta) = \frac{\mathbf{A} \cdot \mathbf{B}}{|\mathbf{A}| |\mathbf{B}|}
\end{equation}

Where $\mathbf{A}$ and $\mathbf{B}$ are $n$-dimensional vectors, $\mathbf{A} \cdot \mathbf{B}$ is the dot product between them, and $|\mathbf{A}| |\mathbf{B}|$ represents the product of their magnitudes. The result of this operation is the angle between these vectors in the embedding space, with values ranging from -1 (indicating opposite vectors) to +1 (indicating proportional vectors). Zero indicates that the vectors are orthogonal. The angle between these vectors in the embedding space can be obtained with $\theta = \arccos(x)$ where $x$ is the output of Equation~\ref{eqn:cos_sim} above.

\begin{figure}[hpbt]
\caption{Cosine Similarity Between Embeddings of Decompressed Text}
\centering
\includegraphics[width=0.45\textwidth]{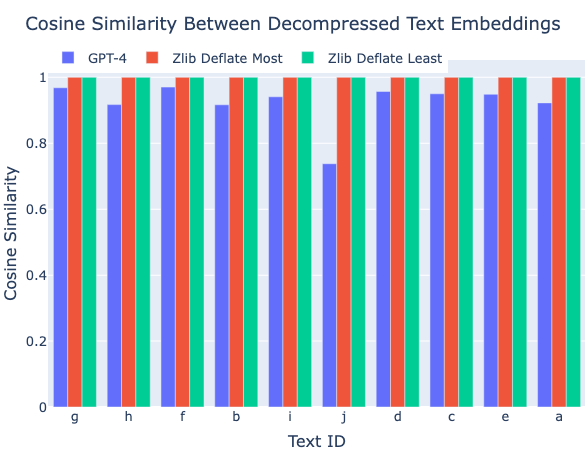}
\label{base_cosine_similarity}
\end{figure}

Figure~\ref{base_cosine_similarity} shows the computed cosine similarity between embedding vectors across all texts for each compression method. 
 \mbox{} \
Again, since Zlib's method is lossless it always captures the semantic meaning since the original input is reconstructed perfectly. GPT-4 does not perfectly preserve the semantic meaning of decompressed texts from the original, but performs relatively well across all texts, with an average angle between embedding vectors of $\arccos(0.923) \approx 22.6^\circ$. 

Interestingly, GPT-4's semantic reconstruction performance is consistent across texts, which is a stark contrast to GPT-4's high volatility in the edit distance metric. From these results, we conclude that GPT-4 may not be suitable for near lossless compression. However, it remains a compelling method to preserve semantic similarity in compressed and decompressed representations, based on the embedding methods we use.

As such, it seems that regardless of whether ChatGPT-4 can accurately capture the underlying text, it is always able to maintain the semantical direction. This aligns much better with LLM's strength of understanding and reasoning with textual relations.

Figure~\ref{base_averaged_cosine_similarity} shows the averaged cosine distance across all texts for each compression method.

As expected, zlib achieves a cosine similarity of 0, which at least indicates proportional magnitude vectors pointed in the same embedding direction. In contrast, ChatGPT-4 produces an average cosine similarity of $0.777$. This result corresponds to an average angular distance between compressed and decompressed representations of roughly 39 degrees. 

Although we can't quantify directly how this angular distance relates to the preservation of underlying semantical meaning in the texts, we can compare the relative performance to the relative performance in edit distance. By taking the relative difference between each model's cosine similarity score, ChatGPT-4 performed, on average, approximately 23\%  worse than zlib's lossless methods for maintaining text semantics. However, when taking the relative difference between edit distances, ChatGPT-4 performed, on average, approximately 33\% worse than zlib's lossless methods for maintaining text character positions. This shows that ChatGPT-4 is relatively better at capturing semantical meaning when compared to maintaining exact text reconstruction.

\subsection{Discussion of Results}
Our findings reported above indicate that GPT-4's compressed text is $3-4\times$ more effective than traditional compression methods. This result, however, is due primarily to GPT-4 discarding half of the originally compressed text during reconstruction, indicated by high edit distances when compared to traditional lossless compression. Despite its inability to perform lossless compression, GPT-4 shows promise in retaining semantic information at a level competitive with lossless compression methods. We quantify semantic preservation of information with the cosine similarity between the embedding of the decompressed text and the embedding of the original text.

Our results suggest that while GPT-4 may not be suitable for applications requiring lossless compression, it may be applicable for use cases where the preservation of semantic information is prioritized. For example, there are several ways to indicate that an event X has occurred in natural language (e.g., ``event X happened'', ``event X occurred'', ``the event called X passed'', etc.). As long as the natural language description semantically captures this single bit of information (whether the event has occurred), then the various text representations indicating this fact are of roughly equal value. 

Motivated by this idea, we applied a similar compression approach for source code. For example, there may be several ways to realize or implement a piece of code to accomplish a task. The quality of a piece of code with respect to its functionality often lies on a spectrum from precision to concision. For example, does the code execute the required functionality? More importantly, what is the minimal representation that achieves the prescribed task that is both performant and secure? Likewise, is there a more accessible path than ever to generate functional source code from natural language using LLMs? In essence, this flexibility in natural language and code expression further motivates our exploration of semantical LLM-based compression.

To summarize, while GPT-4 is not a suitable replacement for traditional lossless compression, it demonstrates potential in preserving the semantic meaning of text. The next two sections explore the results of experiments we conducted to evaluate the influence of prompt structuring on compression quality and code generation from compressed natural language descriptions of code functionality, respectively.

% The analysis of ChatGPT4's compression capabilities highlights the need for further research in the field of LLM-based and approximate compression, as well as the exploration of more effective methods for prompt engineering and compression optimization.

%This aligns with the strengths of large language models in understanding and reasoning about textual relations, though at times in inaccurate or inconsistent ways. The consistency in ChatGPT-4's semantic reconstruction across different texts, despite the variations in edit distance, highlights its ability to maintain the core meaning of the original message. Future research could explore methods to combine the semantic understanding of LLMs like ChatGPT-4 with traditional compression algorithms to achieve a balance between lossless compression and semantic preservation, as well as investigate the factors contributing to ChatGPT-4's varying performance in maintaining character positions.

% \begin{figure}[h!]
% \caption{Averaged Semantical Similarity}
% \centering
% \includegraphics[width=0.5\textwidth]{figures/base_average_cosine.png}
% \label{base_averaged_cosine_similarity}
% \end{figure} \mbox{} \

\section{Prompt Engineering to Facilitate Compression Behaviour}
\label{results_prompt_eng}

This section present the results of our investigation into the role  prompt engineering plays in terms of prompt content and wording in facilitating the compression performance of LLMs, specifically for GPT-4 and GPT-3.5. We examine the impact of three different meta-prompts for compression: \textit{Base Compression} (simply direct to compress the input), \textit{Guided Lossless Compression} (by specifying lossless compression of the input), and \textit{Semantic Compression} (prioritizing semantic recovery).

We found that the choice of meta-prompt influenced the compression behavior of the LLMs studied. To evaluate the effectiveness of compression in relation to edit distance and semantic similarity, we introduced two novel metrics, \textit{Exact Reconstruction Effectiveness} (ERE) and \textit{Semantic Reconstruction Effectiveness} (SRE), respectively. The Exact Reconstruction Effectiveness metric revealed that while the Zlib Deflate lossless compression baselines outperformed GPT-4 and GPT-3.5, our meta-prompts for Guided Lossless Compression method outperformed both Base Compression and Semantic Compression in terms of compression ratio and edit distance minimization.

The Semantic Reconstruction Effectiveness metric, in contrast, demonstrated that the Semantic Compression meta-prompting approach outperformed the baseline lossless compression. Although the baseline lossless methods achieved slightly higher semantic similarity scores, the GPT-4 Semantic Compression model provided an improved compression ratio while preserving functionally equivalent semantic information in the input. This finding suggests that LLM-based Semantic Compression could offer considerable performance and cost gains over traditional compression methods in use cases where the exact reconstruction of the input is not crucial, as long as the underlying meaning remains intact.

%\mbox{} \
\subsection{Experiment Setup}
%\mbox{} \

% Additionally, section \ref{literary_text_performance} will also be used to ground the analysis in concrete external performance through zlib's traditional lossless algorithms.

The aim of this experiment was to distinguish LLM compression behavior and performance when optimized for lossless compression versus semantic compression. We applied the results from Section~\ref{literary_text_performance} to represent the baseline GPT-4 model performance when given no additional specification on compression requirements.  The same analysis was therefore performed as outlined in Section~\ref{literary_text_performance}, but with two separate compression models distinguished by the meta-prompting performed to guide compression behavior. 

GPT-3.5 was also given the same set of prompts over the same text excerpts to compare the compression quality between the different model versions. When requesting lossless compression, each model was fed the following prompt:
\begin{promptbox}
\textit{Please compress the following text into a latent representation that a different GPT-4 model can decompress into the original text. The compression model should be lossless, meaning that a different GPT-4 model should be able to perfectly reconstruct the original text from the compressed representation, without any additional context or information.}
\end{promptbox}
The aim of this prompt was to instruct the model to prioritize lossless compression, thereby ensuring that the decompressed text was as close as possible if not identical to the original input. When decompressing the resulting compressed text, the following prompt was provided to a different GPT-4 model:

\begin{promptbox}
\textit{A different GPT-4 model was given the following prompt: ``Please compress the following text into a latent representation that a different GPT-4 model can decompress into the original text. The compression model should be must be lossless, meaning that a different GPT-4 model should be able to perfectly reconstruct the original text from the compressed representation, without any additional context or information.'' Please decompress the following text into the original text.}
\end{promptbox}
The aim of this prompt was to provide the necessary context for the decompression process, reinforcing the expectation that decompressed text should match the original input exactly. 

When requesting semantic compression, the following prompt was used:
\begin{promptbox}
\textit{Please compress the following text into a latent representation that a different GPT-4 model can decompress into the original text. The compression model should purely minimize the number of characters in the compressed representation, while maintaining the semantics of the original text. The resulting compressed text does not need to be decompressed into the original text, but should capture the semantics of the original text. The compressed text should be able to be decompressed into a text that is semantically similar to the original text, but does not need to be identical.}
\end{promptbox}
The aim of this prompt was to instruct the model to focus on semantic preservation while minimizing the character count, without requiring an exact match between the decompressed and original texts. 

When decompressing the resulting compressed text, the following prompt was used:
\begin{promptbox}
\textit{A different GPT-4 model was given the following prompt: ``Please compress the following text into a latent representation that a different ChatGPT4 model can decompress into the original text. The compression model should purely minimize the number of characters in the compressed representation, while maintaining the semantics of the original text. The resulting compressed text does not need to be decompressed into the original text, but should capture the semantics of the original text. The compressed text should be able to be decompressed into a text that is semantically similar to the original text, but does not need to be identical.'' Please decompress the following text into semantically similar text.}
\end{promptbox}
The aim of this prompt was to provide the necessary context for the decompression process, emphasizing the importance of preserving the semantic meaning of the original text rather than producing an exact match.

GPT-4 was not available programmatically via an API endpoint aat the time we wrote this paper. All prompts and results were therefore obtained manually using the chat interface feature on the ChatGPT website (\texttt{chat.openai.com}). For results from the GPT-3.5 model, the Chat Completion function of the \texttt{openai} Python library was used. This library requires providing the model with a ``system'' prompt to prime the model with expected behavior, as well as the actual chat prompt. For GPT-3.5, the following prompt is used when optimizing for lossless compression:
%\begin{itemize}
 \\ \textbf{System Prompt}:
\begin{promptbox}
\textit{You are a ChatGPT LLM trained by OpenAI to compress text. The compressed text should be able to be decompressed by a different ChatGPT LLM model into the original text. The compression must be lossless, meaning that a different ChatGPT LLM model should be able to perfectly reconstruct the original text from the compressed representation, without any additional context or information. The compressed text does not need to be human-readable, only decompressible by a different ChatGPT LLM model.}
\end{promptbox}

\textbf{Action prompt}:
 \begin{promptbox}
 \textit{Compress the following text. Return only the compressed text with no additional text. Text to compress: ...}
 \end{promptbox}
%\end{itemize}

The following System-Action prompt pair was used to decompress text that was compressed with the previous System-Action prompt pair: \\

\textbf{System Prompt}:
\begin{promptbox}
\textit{You are a ChatGPT LLM trained by OpenAI to decompress text. The compressed text you will be given was compressed by a different ChatGPT LLM that was instructed to perform lossless compression.}
\end{promptbox}

\textbf{Action Prompt}
\begin{promptbox}
    \textit{Decompress the following text. Return only the decompressed text with no additional text. Text to decompress: ...}
\end{promptbox}

When specifying Semantic Compression, the following prompt was used to compress input text: \\ 

\textbf{System Prompt}:
\begin{promptbox}
    \textit{You are a ChatGPT LLM trained by OpenAI to compress text. The compression model should purely minimize the number of characters in the compressed text, while maintaining the semantics of the original text. The resulting compressed text does not need to be decompressed into exactly the original text, but should capture the semantics of the original text. The compressed text should be able to be decompressed into a text that is semantically similar to the original text, but does not need to be identical.}
\end{promptbox}

\textbf{Action Prompt}

\begin{promptbox}
    \textit{Compress the following text. Return only the compressed text with no additional text. Text to compress: ...}
\end{promptbox}

% \begin{itemize}
% \item system: "You are a ChatGPT LLM trained by OpenAI to compress text. The compression model should purely minimize the number of characters in the compressed text, while maintaining the semantics of the original text. The resulting compressed text does not need to be decompressed into exactly the original text, but should capture the semantics of the original text. The compressed text should be able to be decompressed into a text that is semantically similar to the original text, but does not need to be identical."
% \item chat: "Compress the following text. Return only the compressed text with no additional text. Text to compress:"
% \end{itemize}

The following System-Action prompt pair was used for decompressing text  compressed with the previous System-Action prompt pair: \\ 

\textbf{Pre-prompt}

\begin{promptbox}
    \textit{You are a ChatGPT LLM trained by OpenAI to decompress text. The compressed text you will be given was compressed by a different ChatGPT LLM that was instructed to maximize the compression rate and preserve semantical meaning. The decompressed text does not need to match the original exactly, but the decompressed text should have the same semantical meaning as the original text.}
\end{promptbox}

\textbf{Action Prompt:}

\begin{promptbox}
    \textit{Decompress the following text. Return only the decompressed text with no additional text. Text to decompress: ...}
\end{promptbox}
% \begin{itemize}
% \item system: "You are a ChatGPT LLM trained by OpenAI to decompress text. The compressed text you will be given was compressed by a different ChatGPT LLM that was instructed to maximize the compression rate and preserve semantical meaning. The decompressed text does not need to match the original exactly, but the decompressed text should have the same semantical meaning as the original text."
% \item chat: "Decompress the following text. Return only the decompressed text with no additional text. Text to decompress:"
% \end{itemize}

The lossless compression prompts instructed the model to focus on preserving the original text in its entirety, allowing for perfect reconstruction. The lossless compression prompts attempted to accomplish a high-fidelity preservation of the original text, ensuring that no information was lost during the compression and decompression process. The semantic compression prompts instruct the model to prioritize semantic preservation while minimizing the character count. The semantic compression prompts attempted to accomplish a balance between reducing the text size and maintaining semantic integrity.

%\mbox{} \
\subsection{Compressed Text Entropy Analysis}
%\mbox{} \

\begin{figure*}[h!]
\caption{Entropy of Compressed Text For Compression Methods Using Meta-Prompts}
\centering
\includegraphics[width=0.9\textwidth]{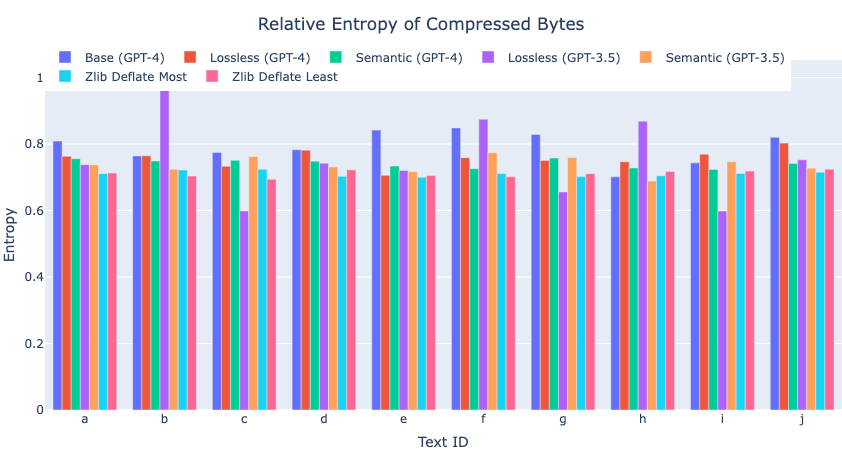}
\label{compression_entropy}
\end{figure*}

Using Equation~\ref{eqn:char_prob} and Equation~\ref{eqn:entropy}, the entropy of the compressed text was computed for each method across all texts. To calculate entropy, the text was first converted into a byte stream representation and the relative probability of characters was then computed. Figure~\ref{compression_entropy} shows the compression entropy for each compression method across all literary texts.
The resulting entropy metrics are fairly similar across all models and texts, with a notable exception being the lossless compression through GPT-3.5. The second column of Table~\ref{tab:avg_eff} gives the averaged compressed text entropy by model.

%Given that ChatGPT-3.5 is not optimized for text compression, a higher level of uncertainty surrounding the generated output is to be expected. 

\begin{table}[hbtp]
\centering
\caption{Average Efficacy Metrics of Compressed Text}
\label{tab:avg_eff}
\begin{tabular}{|c|c|c|c|}
\hline
% CR = compression ratio, ED = edit distance
\textbf{Method} & \textbf{Entropy} & \textbf{CR} & \textbf{ED} \\

%& Avg Cosine Similarity of Embedded Decompression & Relative Exact Effectiveness & Relative Semantic Effectiveness
\hline
Base (GPT-4) & \textbf{0.791} & \textbf{0.825} & 0.510 \\
% & 0.923 & 0.61 & 1\\
\hline
Lossless (GPT-4) & 0.758 & 0.423 & \textbf{0.194} \\
% & 0.976 & 0.945 & 0.542\\
\hline
Lossless (GPT-3.5) & 0.755 & 0.383 & 0.573 \\
%& 0.743 & 0.786 & 0.374\\
\hline
Semantic (GPT-4) & 0.741 & 0.772 & 0.556 \\
% & 0.936 & 0.622 & 0.949\\
\hline
Semantic (GPT-3.5) & 0.737 & 0.768 & 0.556 \\
% & 0.93 & 0.623 & 0.937 \\
\hline
Zlib Deflate Least & 0.711 & 0.453 & 0 \\
% & 1 & 1 & 0.594\\
\hline
Zlib Deflate Most & 0.710 & 0.469 & 0 \\
%& 1 & 0.981 & 0.615\\
\hline
\end{tabular}
\end{table}
Interestingly, the base compression method, where GPT-4 was not explicitly prompted on expected compression behavior, demonstrates the highest entropy. This result was surprising given that ChatGPT-3.5 (which is not specifically designed for this task) performs comparably well. The lossless compression maintains the lowest compressed text entropy, suggesting the efficacy of meta-prompting different models to achieve the desired outcome.

Given the small sample size, however, these results should serve as an initial baseline for subsequent analysis. From these results, we conclude that meta-prompting can concretely impact the compression performance of the LLMs studied. Our findings suggest that GPT-4 is capable of achieving better compression results when given specific instructions about the desired compression behavior, thereby justify further exploration of meta-prompting tactics for improved compression policies.

% This analysis can serve as a starting point for further investigation into the performance of these models and the role of meta-prompting in compression quality optimization.

% \begin{figure*}[h!]
% \caption{Averaged Entropy of Compressed Text}
% \centering
% \includegraphics[width=\textwidth]{figures/averaged_compression_entropy.png}
% \label{averaged_compression_entropy}
% \end{figure*}

%\mbox{} \
\subsection{Compression Ratio}
%\mbox{} \

\begin{figure*}[h!]
\caption{Compression Ratio By Text}
\centering
\includegraphics[width=0.9\textwidth]{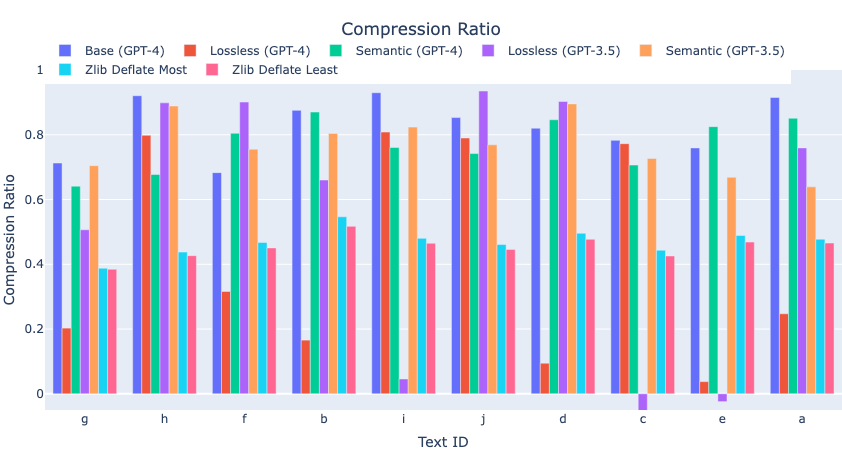}
\label{compression_ratio}
\end{figure*}

Similar to Section~\ref{literary_text_performance}, the compression ratio for each of the candidate compression methods was computed across all texts. Figure~\ref{compression_ratio} plots the derived compression ratio of each method over each text. The graph shows high volatility in model performance across texts and between the models themselves. Interestingly, Lossless GPT-3.5 actually produces a compressed text that is approximately 71\% larger than the original text for text c (the figure is truncated slightly below 0 to save space). Confoundingly, this text's edit distance from the original input text is the worst, whih is another indicator that GPT-3.5 struggles to compress input text, highlighting the limitations of this particular model for compression tasks.

% \mbox{} \

\begin{figure*}[h!]
\caption{Averaged Compression Ratio}
\centering
\includegraphics[width=\textwidth]{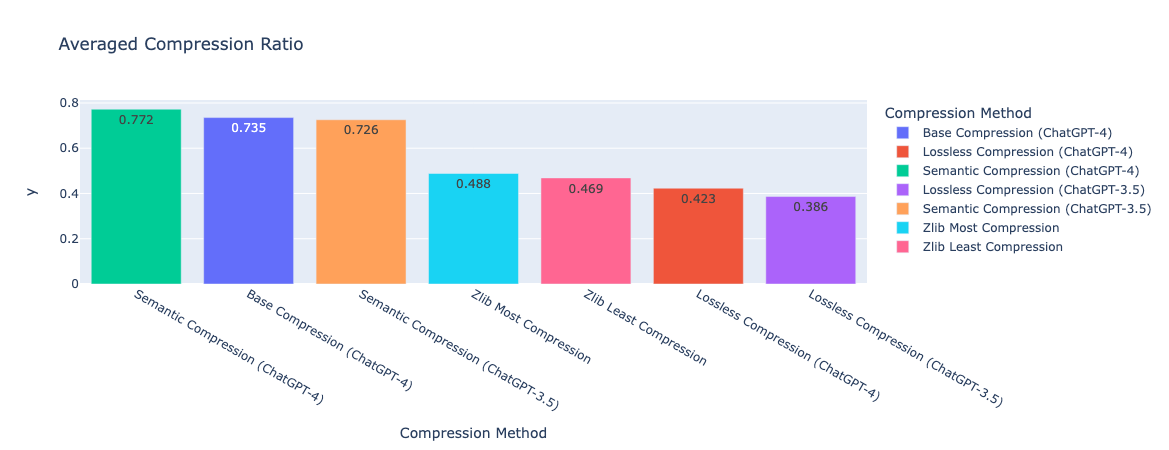}
\label{averaged_compression_ratio}
\end{figure*}% \mbox{} \

The third column of Table~\ref{tab:avg_eff} includes the averaged compression ratio (CR) of each model across all texts. GPT-4 semantic compression maintains the best compression ability, closely followed by the base compression and GPT-3.5's semantic compression. This result was expected since semantic compression is not constrained by reconstructing the exact text and presumably the underlying semantic meaning can be captured in much less text when it can be arbitrarily decompressed. GPT-3.5's competitive semantic compression performance is surprising given its lossless compression performance. One possible explanation is that semantic compression is implicitly derived internally in these LLMs via text embeddings.

The traditional lossless methods sit in the middle of the pack in performance, followed by both GPT lossless methods. The worse compression rates coming from the LLM lossless compression methods is expected and validates the impact of the meta-prompt. Clearly, the models must maintain a greater information density if the intended goal is exact reconstruction. 

Previous LLM exact edit performances were expected to be worse than traditional methods since they maintained significantly less information in the compressed state. As such, they clearly lacked the needed information to reconstruct the text exactly. However, Figure~\ref{averaged_compression_ratio} shows the Lossless LLM compression actually maintain more informational density than traditional lossless methods. 

%Thus, the Lossless LLMs have lost the excuse to not be able to perfectly reconstruct the text. We will unfortunately see them continue to fail, but the model's behavior is directionally correct, if not optimal.

From these results, we conclude that LLMs like GPT-4 can achieve competitive compression rates when given appropriate meta-prompts, particularly in the case of semantic compression. The potential applications of our findings include optimization of data storage and improved communication efficiency. The next steps for this research include investigating the role of fine-tuning LLMs specifically for compression tasks, exploring alternative prompting strategies to further improve compression performance, and examining the impact of larger training data sets on compression outcomes.

%, and better understanding of LLM capabilities in compression tasks. 

%\mbox{} \
\subsection{Edit Distance}
%\mbox{} \

As discussed in Section~\ref{literary_text_performance} the edit distance for each compression method over each text was computed. This metric shows the model's ability to exactly reconstruct the original text from the compressed representation, based on the number of characters that must be inserted or deleted to achieve the original input text from the compressed representation of the text output by the LLM.

The fourth column of Table~\ref{tab:avg_eff} shows the corresponding edit distance for each model over each evaluation text. While the magnitude of edit distance varies greatly between texts, the relative model performance is mostly stable, which suggests that all LLM models struggle on similar kinds of text. Future research should explore the semantic and syntactic classifications of texts that are ``easier'' versus ``harder'' for LLM models to compress effectively. Understanding these relationships enables targeted improvements in model performance and better adaptation to specific use cases.

The most notable model performance deviations are found in the GPT-4 lossless compression. It routinely performs the best of any LLM model, and on texts b (``Break it Down''), d (``Cathedral''), e (``Flowers for Algernon''), and a (``A Good Man is Hard to Find''), the model performs exceptionally well with a normalized edit distance near 0. This result demonstrates that GPT-4 lossless compression can effectively reconstruct text with high accuracy in some cases. We therefore conclude that this model shows promising potential for text compression tasks. 

\begin{figure*}[h!]
\caption{Normalized Decompression Edit Distance by Text}
\centering
\includegraphics[width=0.9\textwidth]{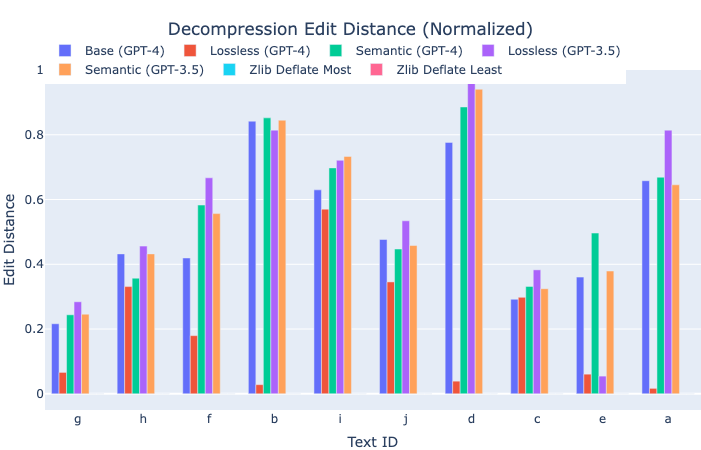}
\label{averaged_edit_distance}
\end{figure*}

Figure~\ref{averaged_edit_distance} averages the model edit distance over all texts. 
As expected, the traditional Zlib lossless compression methods maintain a distance of 0 as they losslessly reconstruct the text. GPT-4 performs better than all remaining models, with over $\approx$50\% more accurate on average than the next closest LLM, GPT-3.5 Semantic Compression. 

Interestingly, GPT-3.5 Semantic Compression and GPT-4 Semantic Compression are roughly split across the texts. Given the other performance differences between the models, this result is unexpected, but is likely justified by the limited evaluation data set, which may not comprehensively capture all model behaviors. This finding also suggests a notion of ``prompt sensitivity'' where the output quality and consistency may vary between different models for similar prompts.

%Given how dependent model performance is on structuring the meta-prompt, it could also be that the used prompt just happened to be better suited for ChatGPT-3.5 rather than ChatGPT-4.

GPT-3.5 Lossless compression clearly underperforms on this evaluation metric. This result aligns with expectations given its subpar performance in both the compression ratio and entropy metrics. Evidently, GPT-3.5 struggles to achieve lossless compression and execute it effectively given the meta-prompt. 

There are several possible explanations for GPT-3.5's limitation. For example, GPT-3.5's training data may not be comprehensive or diverse enough to cover various text compression scenarios, leading to inadequate understanding of lossless compression tasks. Likewise, there may be a difference in model capacity compared to GPT-4, which could result in GPT-3.5 having a reduced ability to distill and generalize compression meta-prompts. 

%Consequently, researchers and practitioners should exercise caution when using ChatGPT-3.5 for lossless compression tasks and explore alternative models, such as ChatGPT-4, to achieve better results.

Figure~\ref{averaged_edit_distance} further validates that the lossless versus semantic meta-prompts in part induce the improved compression behavior in GPT-4. The lossless version performs $\approx$286\% better than Semantic and Base compression for the same model. We therefore conclude from our assessment that meta-prompting for compression has a measurable effect on the quality of text compression tasks.

%approach is effective in controlling the model's behavior for specific tasks. 

% Potential next steps for research include investigating the role of fine-tuning LLMs specifically for compression tasks, exploring alternative prompting strategies to further improve performance, and examining the impact of larger training data sets on compression outcomes.

% \begin{figure*}[h!]
% \caption{Edit Distance by Text}
% \centering
% \includegraphics[width=1\textwidth]{figures/decompression_edit_distance.png}
% \label{edit_distance}
% \end{figure*}

\begin{table}
\centering
\label{tab:avg_qual_decomp}
\caption{Average Quality Metrics of Decompressed Text}
\begin{tabular}{|c|c|c|c|}
\hline
% CS = Cosine Similarity, REE = rel exact eff, RSE = rel sem eff
\textbf{Method} & \textbf{CS} & \textbf{ERE} & \textbf{SRE} \\
\hline
Base (GPT-4) & 0.923 & 0.61 & \textbf{1}\\
%0.791 & 0.825 & 0.510 & 0.923 & 0.61 & 1\\
\hline
Lossless (GPT-4) & \textbf{0.976} & \textbf{0.945} & 0.542\\
% 0.758 & 0.423 & 0.194 & 0.976 & 0.945 & 0.542\\
\hline
Lossless (GPT-3.5) & 0.743 & 0.786 & 0.374\\
%0.755 & 0.383 & 0.573 & 0.743 & 0.786 & 0.374\\
\hline
Semantic (GPT-4) & 0.936 & 0.622 & 0.949\\
% 0.741 & 0.772 & 0.556 & 0.936 & 0.622 & 0.949\\
\hline
Semantic (GPT-3.5) & 0.93 & 0.623 & 0.937 \\
% & 0.737 & 0.768 & 0.556 & 0.93 & 0.623 & 0.937 \\
\hline
Zlib Deflate Least & 1 & 1 & 0.594\\
% 0.711 & 0.453 & 0 & 1 & 1 & 0.594\\
\hline
Zlib Deflate Most & 1 & 0.981 & 0.615\\
% 0.710 & 0.469 & 0 & 1 & 0.981 & 0.615\\
\hline
\end{tabular}
\end{table}

\subsection{Semantic Retention}

As discussed in Section~\ref{literary_text_performance}, the ability of a compression model to retain the underlying semantic meaning was evaluated using the cosine similarity metric between the embedding vectors from the original and decompressed text. Cosine similarity measures the angle between two embedding vectors, effectively capturing the degree of similarity between the original and decompressed text in the embedding space.

\begin{figure*}[h!]
\caption{Cosine Similarity by Text}
\centering
\includegraphics[width=0.9\textwidth]{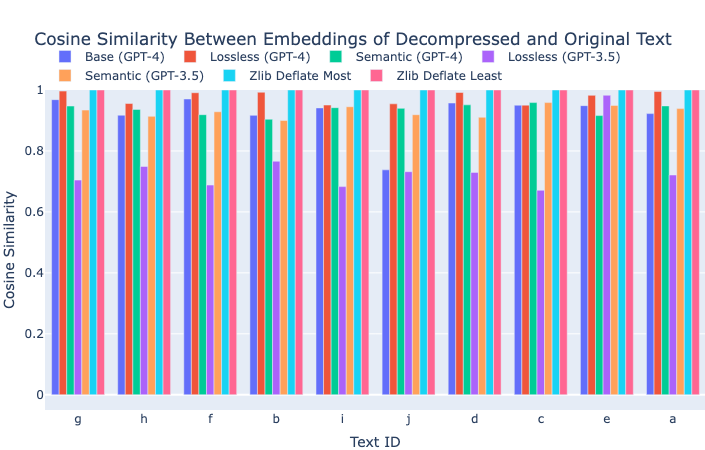}
\label{cosine_sim}
\end{figure*}

Figure~\ref{cosine_sim} plots the resultant cosine similarity between embeddings of the decompressed and original texts for each model as an indicator of semantic retention from input text to output compressed text. 
 % \mbox{} \
Both the overall magnitude and relative performance difference between models remain fairly stable across texts, which demonstrates the innate ability of LLMs to capture the underlying semantic context at a much higher level than exact text reconstruction. This behavior occurs because LLMs are designed to understand and generate meaningful text, so their internal representations inherently encode semantic information.

Again, the Zlib baseline methods using the Deflate algorithm exhibit perfect semantic retention since they are lossless. We also observe that the GPT-4 Lossless compression approach achieves perfect semantic retention in several cases. GPT-4 Semantic compression performs nearly as well, indicating that it can effectively retain the semantic content of the text even when the exact reconstruction is not required. This result highlights the potential of using LLMs for semantic compression tasks in cases where the exact reconstruction of the input data is not strictly necessary.

% \begin{figure*}[h!]
% \caption{Averaged Edit Distance by Text}
% \centering
% \includegraphics[width=\textwidth]{figures/averaged_decompression_cosine_sim.png}
% \label{averaged_cosine_sim}
% \end{figure*} \mbox{} \

The second column of Table~\ref{tab:avg_qual_decomp} gives the average cosine similarity (CS) of the models across all texts. GPT-4 Lossless is the strongest performing LLM model. GPT-4 Semantic Compression follows closely behind, with a 4\% drop in performance. 

Surprisingly, GPT-3.5 Semantic Compression nearly perfectly matches the GPT-4 Semantic Compression performance, to the point where the exact ordering is likely dependent on internal model randomness. GPT-4 Base Compression and GPT-3.5 Lossless Compression perform considerably worse. These results reinforce that meta-prompting contributes at least in part to the overall performance in semantic retention of compressed text. 

%This is important because it showcases the potential for using LLMs in applications that require compression while preserving the semantic meaning of the text, such as summarization, data storage, and transmission.

%\mbox{} \
\subsection{Exact Reconstruction Effectiveness}
%\mbox{} \

Exact Reconstruction Effectiveness is a novel metric proposed in this paper to capture compression performance with respect to both the compressed size and the ability to perfectly reconstruct the original text. This metric provides a fair means to compare LLM compression performance with traditional lossless algorithms. While traditional lossless algorithms can perfectly reconstruct the text, their resulting compression rate is often much lower than LLM compression. 

In contrast, LLM compression derives a more efficient compressed representation, though it struggles to perfectly replicate the original input text. Exact Reconstruction Effectiveness reconciles these differences to provide a balanced evaluation metric. Exact Reconstruction Effectiveness is computed by taking the inverse of the log-normalized compression ratio multiplied by the inverse of the edit distance, as shown in the following equation:

\mbox{} \
\begin{equation}
\begin{aligned}% [b]
\textit{Exact Reconstruction Effectiveness} \\ = \frac{1}{\log(\textit{Compression Ratio})} \times \frac{1}{\textit{Edit Distance}}
\end{aligned} %[b]
\end{equation}
\mbox{} \
This equation maximizes the compression ratio and minimizes the edit distance (or in this case, maximizes the inverse of the edit distance as this accounts for numerical instability caused by 0 values in the edit distance). Taking the logarithm of the compressed ratio scales the relative importance of more effective methods. This metric is concerned with the ability to reconstruct the exact input, so we do not want to allow methods that have a disproportionately high compression rate, but terrible edit distance, to score highly. 

For example, if a method discards 99\% of the text, its compression rate alone could be enough to compensate for a very high edit distance. Taking the logarithm of the compression ratio mitigates the impact of such situations and balances between both the compression ratio and the edit distance. We take the inverse of the entire equation as the compression rates are strictly less than 1, resulting in negative values when log-normalized. Inverting this value is not mathematically necessary, but maximizes a more intuitive positive number, rather than minimizing a negative number.

% \begin{figure*}[h!]
% \caption{Exact Reconstruction Effectiveness}
% \centering
% \includegraphics[width=1\textwidth]{figures/exact_compression_effectiveness.png}
% \label{exact_compression_effectiveness}
% \end{figure*} \mbox{} \

The third column of Table~\ref{tab:avg_qual_decomp} shows the Normalized Exact Reconstruction Effectiveness for each model. Zlib's least compression method scores marginally higher than the most method due to the inherent trade-off in compression ratio. GPT-4 lossless compression scores second highest and validates both the metric formulation and the meta-prompting technique. GPT-4 semantic compression scores the worst as it is optimized for the opposite use case and thus its added benefit of best-in-class compression rate is mitigated and its average performance in exact reconstruction is unable to compensate. 

%GPT-3.5 lossless compression is the next tier in performance, landing solidly in the middle of the pack, before being followed by ChatGPT-3.5 Semantic compression with a corresponding sharp drop in performance.

From these results, we conclude that when accounting for the added benefit of smaller compression sizes, correctly prompted LLM models can compete with state-of-the-art traditional lossless compression methods for certain use cases. 

The potential applications of our findings include efficient data storage and retrieval, particularly in situations where exact text reproduction is not strictly necessary, and maintaining the semantic information is of greater importance than preserving the exact structure of the text. Moreover, this research opens up new avenues for exploring LLMs and their potential in other domains where compression is essential, such as multimedia and sensor data. Future work could investigate the development of domain-specific prompts and techniques to further optimize LLM-based compression for a wider range of applications.

%Naturally, the lossless LLM models cannot be used for tasks that require a reconstruction performance guarantee, but they are effective for situations in which the underlying text structure needs to be preserved, but a missing comma has a negligible impact.

%As such, we can conclude that LLMs, when appropriately prompted, can provide a competitive alternative to traditional lossless compression methods in certain use cases. 

% In summary, our findings demonstrate that LLMs, when properly guided through meta-prompting, can achieve competitive compression performance compared to traditional lossless algorithms. This not only highlights the versatility and adaptability of LLMs but also paves the way for the integration of these models into a broader spectrum of compression tasks and applications.

%\mbox{} \
\subsection{Semantic Reconstruction Effectiveness}
%\mbox{} \

Semantic Reconstruction Effectiveness is the other novel metric proposed in this paper. We use this metric to evaluate the performance of compression algorithms with respect to the captured underlying semantic context. This comparison is particularly useful for scenarios where exact text reconstruction is not a strict requirement, and the focus is instead on preserving the overall meaning of the original text.

\mbox{} \
\begin{equation}
\begin{aligned}
\textit{Semantic Reconstruction Effectiveness}  \\ = \text{Compression Rate} \times \text{Cosine Similarity}
\end{aligned}
\end{equation}
% \mbox{} \

% \begin{figure*}[h!]
% \caption{Semantical Reconstruction Effectiveness}
% \centering
% \includegraphics[width=1\textwidth]{figures/semantic_comp_eff.png}
% \label{semantic_compression_effectiveness}
% \end{figure*} % \mbox{} \

The fourth column of Table~\ref{tab:avg_qual_decomp} displays the computed Semantic Reconstruction Effectiveness for each model. The results validate our metric approach and the effectiveness of the meta-prompting technique, as GPT-4 Semantic Compression emerges as the highest-performing model under this metric by a considerable margin. In contrast, the two lossless compression models exhibit the worst performance since they are optimized for the exact opposite use case, which minimizes the data size at the expense of discarding semantically rich information from the input text. Interestingly, traditional lossless methods demonstrate average performance under this metric.

% Moreover, GPT-3.5 Semantic Compression maintains the second-best performance, lending further credibility to our conclusions.

These findings again highlight the potential of LLMs such as GPT-4 in semantic compression applications where exact text representation is not a priority. For example, in a scenario where an LLM generates a tailored sales pitch using a context document, maintaining the precise wording or structure of the context document is not crucial, as long as the core semantic selling points are retained. With Semantic Compression, the input prompt size can be reduced by nearly 80\%, yielding functionally equivalent output while substantially increasing the information density that can be fed to the model and simultaneously reducing operating costs.

%Another example use case for semantic compression involves enabling hyper-communication between LLMs, which can have far-reaching implications in the field of artificial intelligence.

%Consider a scenario where multiple LLMs are working together to solve complex problems or complete intricate tasks that require collaborative efforts. In such cases, these models need to exchange information with each other in an efficient and effective manner. By employing semantic compression, the LLMs can greatly reduce the size of the exchanged messages while preserving the core semantic information. This allows for faster communication, lower bandwidth consumption, and potentially reduced latency, enabling the LLMs to collaborate more effectively and accomplish tasks more rapidly.

%This hyper-communication can be particularly beneficial in distributed AI systems, where LLMs might be geographically dispersed or operating under resource constraints. For instance, in a multi-agent system that relies on collaborative learning, semantic compression could significantly enhance communication efficiency between agents, accelerating their learning process and leading to better overall performance.

Moreover, the ability to exchange compressed---but semantically rich---information can open up new possibilities in the development of AI systems with modular architectures, e.g., where specialized LLMs can communicate with each other to form more powerful and adaptable problem-solving systems. This design can potentially yield AI systems that can tackle a broader range of challenges, as well as allow more efficient transfer learning between models. 

To summarize, we demonstrate the value of our novel metric, Semantic Reconstruction Effectiveness, in assessing the performance of compression algorithms with respect to captured semantic context between input and output text. The superior performance of GPT-4 Semantic Compression in this regard, as well as the potential applications of semantic compression in various use cases, underscores the versatility and adaptability of LLMs for tasks where preserving meaning is more important than exact text reconstruction.

%Semantic compression can play a crucial role in enabling hyper-communication between LLMs, paving the way for more efficient, flexible, and collaborative AI systems. The potential implications of this development include faster learning, enhanced performance, and the creation of more versatile AI architectures that can address a wide range of complex problems.

\lstset{
    basicstyle=\ttfamily\small,
    breaklines=true,
    postbreak=\mbox{\textcolor{red}{$\hookrightarrow$}\space},
    linewidth=\columnwidth % Set the maximum width to the column width
}

\section{Semantic Compression For Code Generation}
\label{results_code_integration}

This section presents the results of our third experiment, which investigated the application of semantic compression to code generation. The objective of this experiment was to evaluate the potential of using descriptions of code compressed by GPT-4 to reconstruct source code and determine functional equivalence.

Our experiment began with GPT4-generated descriptions for a set of Python functions. These descriptions were then used as inputs to a separate model to regenerate the source code of these functions, establishing a baseline for comparison. Next, the descriptions were semantically compressed and provided to another model instance (i.e., a separate GPT-4 conversation) for code reconstruction. Finally, GPT-4 was used to determine whether the reconstructed functions were functionally equivalent to the original ones.

The results from our experiments demonstrated that GPT-4 performed surprisingly well in this context. Four of the six functions (including examples from the \texttt{pandas} library) were reconstructed perfectly, with two functions slightly incorrect in handling individual characters instead of string instances for an input list of strings. This result suggests that semantic compression for code generation tasks can effectively retain the necessary information for code reconstruction and warrants further exploration to enhance the efficiency and cost-effectivness of LLM-based code generation and summarization, and to better understand the limitations of LLMs in manipulating code or text obtained from compressed prompts.

% These findings emphasize the potential of LLM semantic compression for enhancing the efficiency and cost-effectiveness of code generation tasks. Further research is needed to explore the integration of semantic compression with other types of textual data, as well as the impact of nested or recursive compression on information loss.

\subsection{Experiment Setup}

To evaluate GPT-4's ability to use semantic compression for code generation, we create six Python functions of increasing complexity. The \texttt{count\_string\_instances} function takes a list of strings as input and returns a dictionary containing the count of each string instance in the input list:

\mbox{}
\begin{lstlisting}[language=Python]
def count_string_instances(input: List[str]) -> Dict[str, int]:
    return { s: input.count(s) for s in input}
\end{lstlisting}
\mbox{}

The \texttt{duplicate\_strings} function takes a list of strings as input and returns a dictionary with each string as a key and the value as the key string concatenated to itself, repeated twice:

\mbox{}
\begin{lstlisting}[language=Python]
def duplicate_strings(input: List[str]) -> Dict[str, str]:
    return { s: s + s for s in input}
\end{lstlisting}
\mbox{}

The \texttt{increment\_value\_at\_string} function takes a dictionary with string keys and integer values as input and returns a new dictionary with the same keys but with all values incremented by 1:

\mbox{}
\begin{lstlisting}[language=Python]
def increment_value_at_string(input: Dict[str, int]) -> Dict[str, int]:
    return { key: value + 1 for key, value in input.items()}
\end{lstlisting}
\mbox{} \

The \texttt{return\_one} function accept any number of positional and keyword arguments and returns the integer 1:

\mbox{} \
\begin{lstlisting}[language=Python]
def return_one(*args, **kwargs) -> int:
    return 1
\end{lstlisting}
\mbox{} \

The \texttt{append\_mutated} function takes a list of strings as input and mutates each string in the input list by appending the string ``mutated'':

\mbox{} \
\begin{lstlisting}[language=Python]
def append_mutated(input: List[str]) -> None:
    for i in range(len(input)):
        input[i] = input[i] + 'mutated'
\end{lstlisting}
\mbox{} \

Finally, the \texttt{datetime\_to\_prob} function takes a Pandas Series object that is called \texttt{created\_at\_dt} containing \texttt{datetime} objects and a frequency string \texttt{freq}: 

\mbox{} \
\begin{lstlisting}[language=Python]
def datetime_to_prob(created_at_dt: pd.Series, freq: str) -> Dict[str, int]:
    # Group by day
    created_at_freq = pd.to_datetime(created_at_dt.apply(lambda x: x.strftime(freq)))
    created_at_freq_grouped: List[Tuple[pd.Timestamp, pd.Series]] = created_at_freq.groupby(created_at_freq)
    # Get the day to frequency of account creation mapping
    day_to_creation_freq = {}
    for group_num, group in created_at_freq_grouped:
        day_to_creation_freq[group_num] = len(group)
    
    # Normalize the frequencies
    for day in day_to_creation_freq:
        day_to_creation_freq[day] /= len(created_at_freq)
    
    # Create rolling cumulative sum
    day_to_creation_freq = {day: sum(day_to_creation_freq[day_] for day_ in day_to_creation_freq if day_ <= day) for day in day_to_creation_freq}

    # Create rolling probability
    for day in day_to_creation_freq:
        day_to_creation_freq[day] /= day_to_creation_freq[max(day_to_creation_freq)]

    return day_to_creation_freq
\end{lstlisting}
\mbox{} \

As shown above, the \texttt{datetime\_to\_prob} function groups the \texttt{datetime} objects by the specified frequency, counts the occurrences of each group, normalizes the frequencies, computes the rolling cumulative sum, and calculates the rolling probability of each group. This function returns a dictionary mapping each \texttt{datetime} group to its corresponding rolling probability.

Four instances of GPT-4 were then applied, as follows. A first GPT-4 instance was fed the source code of each function shown above and asked to generate a text description of the function's behavior. A second GPT-4 instance reconstructed the Python function source code using the unaltered text description of the original code to establish a baseline. We then employed a third GPT-4 instance and asked it to semantically compress the text description of the function of interest. Finally, the compressed text description was given to a fourth GPT-4 instance to reconstruct the Python function source code for evaluation.

\subsection{Evaluating GPT-4 Semantic Compression}
 
To evaluate the ability of GPT-4 to use semantic compression for code generation, the first GPT-4 instance was fed each of the above example functions with the following prompt:

\begin{promptbox}
\textit{Given the following python function, provide a description of its expected behaviour. This description should be in plain English and should be as detailed as possible. It should be able to be used by a separate GPT-4 model to reconstruct the function.}
\end{promptbox}

The model generated a detailed description of each function's expected behavior. Subsequently, the second GPT-4 instance was given the following prompt to semantically compress the function descriptions generated by the first GPT-4 instance:

\begin{promptbox}
\textit{Compress the following python function description into its smallest possible representation. The resulting compressed text does not need to be human readable, but should be able to be used by a separate GPT-4 model to reconstruct the function description.}
\end{promptbox}

The model produced compressed text descriptions for each function. To establish a baseline, the third GPT-4 instance was given the following prompt to reconstruct the Python function using the unaltered text description:

\begin{promptbox}
\textit{A different GPT-4 model gave the following description of a python function. Reconstruct the function from the following description:}
\end{promptbox}

Finally, the fourth GPT-4 instance was given the following prompt to reconstruct the Python function using the compressed text description:

\begin{promptbox}
\textit{A different GPT-4 model gave the following compressed description of a python function. Reconstruct the function from the following description: ...}
\end{promptbox}

We then evaluated GPT-4's ability to leverage semantic compression for code generation tasks by comparing the reconstructed functions using both the unaltered and semantically compressed text descriptions, as discussed next.

%This experiment provided insights into the potential applications of semantic compression in improving the information density and reducing the operating costs of working with large language models.

\subsection{Code Generation from Base Descriptions}

Establishing a performance baseline was crucial to understand the effectiveness of the semantic compression in the code generation task. It allowed us to compare the quality of the reconstructed functions using both the unaltered and compressed text descriptions. This comparison was important because it provided insights into how well the GPT-4 model leveraged semantic compression to generate both accurate and concise code. 

Existing research in this domain has already been conducted~\cite{chen2021evaluating}, demonstrating the potential of LLM usage in code generation tasks. Given the supplied base function descriptions, an independent GPT-4 model generated the  functions discussed below.

The generated \texttt{chatgpt\_base\_gen\_function\_one} function is partially correct because it counts the occurrences of each character in the input list of strings and returns a dictionary with the character count instead of the string instance count as the original function:

\begin{lstlisting}[language=Python]
def chatgpt_base_gen_function_one(strings_list):
    char_count = {}
    for string in strings_list:
        for char in string:
            if char in char_count:
                char_count[char] += 1
            else:
                char_count[char] = 1
    return char_count
\end{lstlisting}

The generated \texttt{chatgpt\_base\_gen\_function\_two} above is also partially correct because it identifies all unique characters from the input list of strings and returns a dictionary with each character mapped to its doubled version, instead of mapping each input string in the list to a duplicated version of itself and keyed by the original string:

\begin{lstlisting}[language=Python]
def chatgpt_base_gen_function_two(strings_list):
    unique_chars = set(''.join(strings_list))
    doubled_chars = {char: char * 2 for char in unique_chars}
    return doubled_chars
\end{lstlisting}

The generated \texttt{chatgpt\_base\_gen\_function\_three} code is correct because it takes an input dictionary and returns a new dictionary with the values incremented by one:

\begin{lstlisting}[language=Python]
def chatgpt_base_gen_function_three(input_dict):
    incremented_values = {key: value + 1 for key, value in input_dict.items()}
    return incremented_values
\end{lstlisting}

The \texttt{chatgpt\_base\_gen\_function\_four} code is correct because it accepts any number of positional and keyword arguments and simply returns the integer 1:

\begin{lstlisting}[language=Python]
def chatgpt_base_gen_function_four(*args, **kwargs):
    return 1
\end{lstlisting}

The generated \texttt{chatgpt\_base\_gen\_function\_five} is correct because it appends the string ``mutated'' to each element in the input list of strings:

\begin{lstlisting}[language=Python]
def chatgpt_base_gen_function_five(strings_list):
    for i in range(len(strings_list)):
        strings_list[i] += 'mutated'
\end{lstlisting}

The generated \texttt{chatgpt\_base\_gen\_function\_six} code is correct because it processes a pandas Series of datetime objects, groups them by the given frequency, computes normalized frequencies, and returns a dictionary with the rolling probability of account creation:

\begin{lstlisting}[language=Python]
def chatgpt_base_gen_function_six(created_at_dt, freq):
    # Step 1: Convert to freq format
    created_at_freq = created_at_dt.dt.strftime(freq).map(pd.Timestamp)

    # Step 2: Group by unique Timestamp values
    created_at_freq_grouped = list(created_at_freq.groupby(created_at_freq).groups.items())

    # Step 3: Initialize the empty dictionary
    day_to_creation_freq = {}

    # Step 4: Iterate through the grouped list and populate the dictionary
    for group_num, group in created_at_freq_grouped:
        day_to_creation_freq[group_num] = len(group)

    # Step 5: Normalize the frequencies
    total_length = len(created_at_freq)
    day_to_creation_freq = {key: value / total_length for key, value in day_to_creation_freq.items()}

    # Step 6: Update the dictionary with a rolling cumulative sum
    rolling_sum = 0
    for key in day_to_creation_freq:
        rolling_sum += day_to_creation_freq[key]
        day_to_creation_freq[key] = rolling_sum

    # Step 7: Create a rolling probability
    max_value = max(day_to_creation_freq.values())
    day_to_creation_freq = {key: value / max_value for key, value in day_to_creation_freq.items()}

    # Step 8: Return the dictionary
    return day_to_creation_freq
\end{lstlisting}

From these results, we conclude that GPT-4 is capable of near-perfect reconstruction of the original functions using unaltered text descriptions. GPT-4 was confused by the first two of these test functions, operating at the character level instead of the string level for the inputs lists. These results provide a promising baseline against which to compare LLM performance when reconstructing functions from compressed text descriptions.

\subsection{Code Generation from Compressed Text}

Now that we established GPT-4 can correctly reconstruct functions from text descriptions, we explored if it can generate the functions equally well when given the text description in compressed form. As shown in the previous experiments, we observed substantial exact text loss when using LLM compression. Our next experiment aimed to explore if semantic retention could overcome information loss due to compression when writing source code. Given the compressed description prompt, GPT-4 generated the reconstructed Python functions discussed below.

The generated \texttt{chatgpt\_base\_gen\_function\_one} code is partially correct because it counts the occurrences of each unique character in the input list of strings and returns a dictionary with the character count, instead of the count of string instances:

\begin{lstlisting}[language=Python]
def chatgpt_base_gen_function_one(l: list[str]) -> dict:
    unique_chars = set(''.join(l))
    occurrences = {k: sum(s.count(k) for s in l) for k in unique_chars}
    return occurrences
\end{lstlisting}

The generated \texttt{chatgpt\_base\_gen\_function\_two} code is also partially correct because it identifies unique characters from the input list of strings and returns a dictionary with each character mapped to its doubled version, instead of duplicating the input string:

\begin{lstlisting}[language=Python]
def chatgpt_base_gen_function_two(l: list[str]) -> dict:
    unique_chars = set(''.join(l))
    doubled_chars = {k: k * 2 for k in unique_chars}
    return doubled_chars
\end{lstlisting}

The generated \texttt{chatgpt\_base\_gen\_function\_three} is correct because it takes an input dictionary and returns a new dictionary with the values incremented by one:

\begin{lstlisting}[language=Python]
def chatgpt_base_gen_function_three(d: dict[str, int]) -> dict:
    incremented_values = {k: v + 1 for k, v in d.items()}
    return incremented_values
\end{lstlisting}

The generated \texttt{chatgpt\_base\_gen\_function\_four} is correct because it accepts any number of positional and keyword arguments and simply returns the integer 1:

\begin{lstlisting}[language=Python]
def chatgpt_base_gen_function_four(*args, **kwargs) -> int:
    return 1
\end{lstlisting}

The generated \texttt{chatgpt\_base\_gen\_function\_five} code is correct because it appends the string 'mutated' to each element in the input list of strings:

\begin{lstlisting}[language=Python]
def chatgpt_base_gen_function_five(l):
    for i in range(len(l)):
        l[i] += 'mutated'
    return None
\end{lstlisting}

The generated \texttt{chatgpt\_base\_gen\_function\_six} code is correct because it processes a Pandas Series of datetime objects, groups them by the given frequency, computes normalized frequencies, and returns a dictionary with the rolling probability of account creation:

\begin{lstlisting}[language=Python]
def chatgpt_base_gen_function_six(created_at_dt, freq):
    # Step 1: Convert created_at_dt to created_at_freq using the given frequency
    created_at_freq = created_at_dt.resample(freq).count()

    # Step 2: Group created_at_freq into created_at_freq_grouped
    created_at_freq_grouped = list(created_at_freq.items())

    # Step 3: Initialize day_to_creation_freq dictionary
    day_to_creation_freq = {}

    # Step 4: Iterate through created_at_freq_grouped and update day_to_creation_freq
    for grp_num, grp in created_at_freq_grouped:
        day_to_creation_freq[grp_num] = len(grp)

    # Step 5: Normalize day_to_creation_freq
    total = sum(day_to_creation_freq.values())
    day_to_creation_freq = {k: v / total for k, v in day_to_creation_freq.items()}

    # Step 6: Update day_to_creation_freq with rolling cumulative sum
    rolling_cumsum = pd.Series(day_to_creation_freq).cumsum().to_dict()
    day_to_creation_freq.update(rolling_cumsum)

    # Step 7: Create rolling_prob by dividing each value by the maximum value
    max_val = max(day_to_creation_freq.values())
    day_to_creation_freq = {k: v / max_val for k, v in day_to_creation_freq.items()}

    # Step 8: Return day_to_creation_freq
    return day_to_creation_freq
\end{lstlisting}

From these initial experiments, we conclude that GPT-4 can accurately reconstruct the original Python functions from semantically compressed text descriptions of Python source code. This result indicates that despite the information loss due to compression, the model can retain the essential semantics required for code generation. This finding also demonstrates the potential of using semantic compression for efficient communication and processing of source code by LLMs.

\subsection{Result Implications}

Our experiments demonstrate that GPT-4 can reliably reconstruct code when given a compressed text description, which suggests it can successfully retain essential semantic information despite information loss due to compression, enabling it to reconstruct the original code with functional accuracy.
Assuming the prompt size needed to generate and broadly understand basic source code functionality can be decrease substantially, the throughput of LLMs becomes considerably larger. Given an average semantic compression rate of approximately 80\%, this brings ChatGPT-4's effective token limit from around $\approx32$k tokens to:

\mbox{} \
\begin{equation}
\begin{aligned} %[b]
\textit{Effective Token Limit}  = \\
32,000 \times \frac{1}{1 - 0.80} = 160,000
\end{aligned} %[b]
\end{equation}
\mbox{} \

This $\sim$$5\times$ expansion in tokens increases the effective capacity of the model to handle larger tasks related to code summarization, annotation, and generation. This finding is important because it enables researchers and practitioners to use LLMs like GPT-4 for potentially larger and more intricate code generation and interpretation tasks that were previously infeasible due to token limitations restricting the total prompt size.

We conclude from our experiments that semantic compression is a powerful technique for improving the efficiency and effectiveness of LLMs in code generation and interpretation tasks. One possible improvement to our approach involves strictly enforcing the use of type hints when prompting for code generation to further provide an LLM context about expected variable types and their associated transformations.

%By leveraging compressed text embeddings, we can harvest additional functionality from ChatGPT-4, unlocking new possibilities for their application in various domains, including software engineering, artificial intelligence, and beyond.

\section{Additional Exploration}
\label{additional}

This section outlines additional studies that were conducted as part of this investigation, along with a brief discussion of our initial findings.

% \begin{itemize}
%     \item Compression of factual information from Wikipedia
%     \item Evaluating the capabilities of LLMs to manipulate and maintain compressed information over time
%     \item Meta-programming of code based on terse descriptions of code
% \end{itemize}

\subsection{Compression of Factual Information}

In addition to literary short stories, the compression tasks of Section~\ref{literary_text_performance} were also conducted on nine randomly selected Wikipedia pages on the following topics: basketball, the history of medicine, quantum mechanics, the Roman empire, R2K, the solar system, tulip (flower), viola (instrument), WeWork (company). The results from this task were similar to the results reported in Section~\ref{literary_text_performance} and are therefore omitted for brevity.

\subsection{Manipulation of Compressed Information}

As an initial foray into effective manipulation of compressed representations, we asked GPT-4 in separate tasks to recursively (1) form a short story with single sentences, starting from the single sentence seed \textit{``A small furry dog with a blue collar visited a big city by itself.''} and (2) construct a list of digits from 0-9, adding one digit at a time, using the number \texttt{5} as a starting point. Both tasks were subject to a maximum length constraint, so the model could determine the point of task completion. 

Directives were provided in a model-compressed format, so the model was expected to first decompress the directive, apply the instructions of the decompressed directive to the compressed data, and recompress the modified data with the original data, to pass to another GPT-4 instance for further processing. The model did not perform well on either of these tasks since it required recursive manipulation and iterative compression of information. We will explore this direction further in future work.

\subsection{Compressive Meta-Prompting for Code Generation}

One view of compression from the perspective of source code generation is the length of a minimal text description of source code that encodes a functional code prototype. To further explore the code generation task, a meta-prompt was provided to GPT-4 to enable a code generation directive and a language change directive. 

These directives were encoded as follows: \texttt{cg (lang) [lib] (func)} enables ``code generation'' in a specific language, possibly using a specific library, implementing the described functionality; \texttt{chg (new\_lang)} enables a change from the previously generated code language to the target language. In this context parentheses indicate mandatory arguments and brackets indicate optional arguments.

In limited experiments, this compressive meta-prompt facilitates the implementation of a basic recurrent neural network (RNN) in Python, using PyTorch, with the command \texttt{cg Python torch rnn}. This code was then converted to Rust source code using \texttt{chg Rust}. In a separate experiment, a semi-colon was used without prior instruction to produce Python code using NumPy for matrix multiplication and then generate code with the same functionality in JavaScript using the chained directives: \texttt{cg Python numpy matrix multiplication; chg JavaScript; cg JavaScript matrix multiplication}. 

These results from this experiment warrant additional exploration in what we call ``compressive meta-prompts.'' These types of meta-prompts are directives that encode a potentially high information density for a sufficiently specified objective.

\section{Related Work}
\label{related}

This section compares our approach with related work on evaluating LLMs and data compression.

\subsection{Neural Data Compression}

Data compression aims to reduce the size of data in a way that maximally preserves the original raw data before compression. This problem has recently been addressed via neural models \cite{yang2022introduction, alwani2022decore} which can achieve more nuanced compression policies than rigidly defined algorithms or encoding schemes. These approaches may offer promise for complex compression tasks, such as for high-entropy data or intricately structured data formats, at the expense of increased performance overhead or limited generalization power over legacy approaches. 

Our work leverages LLMs, which are treated as a black box with minimal interpretability from input prompt to output response. Our approach is based on the view that the model weights of an LLM represent a compressed representation of its training data and can thus serve as a compression mechanism for input data via strategic prompting. Motivated by fundamental constraints on input token counts \cite{openai_token_limits} and request counts to an LLM, we explore the compression capabilities of LLMs in both lossless and lossy paradigms on both text-to-text and text-to-code tasks. While our results are not overwhelmingly impressive relative to legacy lossless techniques, they warrant additional exploration of meta-prompting for semantic compression capabilities.

\subsection{Large Language Model Code Generation}

The code generation capabilities of LLMs have marked a paradigm shift in legacy software engineering workflows. Tools like Github Co-Pilot and similar offerings \cite{github, xia2023conversational} provide in-IDE pair programming capabilities, endowing programmers with auto-complete suggestions for code modification and refactoring, based on comments or existing neighboring code. These tools are sometimes viewed as simple shortcuts to crowd-sourcing sites, such as StackOverflow. Indeed, studies~\cite{dakhel2022github, pearce2021asleep, asare2022github, sandoval2023lost, chen2021evaluating} have shown their propensity to propagate low-quality or insecure code by de-sensitizing  programmers to the quality of generated code. 

Our work focuses on preliminary investigations of leveraging LLMs to compress natural language descriptions of code functionality. We also focus on generating source code from compressed, LLM-generated natural language descriptions of Python functions.

\subsection{Large Language Model Evaluation}

Empirical evaluation of LLMs is still a nascent discipline. In particular, there is both interest and urgency in improving our collective understanding of the ways an LLM produces an output response given an input prompt to (1) enhance transparency and confidence around LLM applications and (2) mitigate systematic bias \cite{liang2022holistic, meister2021language, 10.1162/coli_a_00355, bubeck2023sparks, bommasani2021opportunities, bang2023multitask}. Efforts in this area center on the global statistical trends in natural language usage as it applies to prompting and downstream use of output in LLMs. 

There are a number of open questions with respect to fair evaluation of LLMs and mitigation of bias in these models, as well as the most productive techniques for ensuring high quality responses from an input prompt \cite{liu2023pre, white2023prompt}. 
Our work assesses the affects of prompt engineering on the measurable effects of compression quality for text compression and code generation and summarization tasks. Evaluation methods of LLMs stand to further enhance and contextualize the findings in this work, but much additional research is required in this area.

\section{Concluding Remarks}
\label{conclusion}

This paper presents an initial evaluation of compression techniques in Large Language Models (LLMs), specifically ChatGPT-3.5 and ChatGPT-4. We propose two novel metrics for evaluating their performance: Semantic Reconstruction Effectiveness (SRE) and Exact Reconstruction Effectiveness (ERE). We also demonstrate that LLMs can effectively reconstruct source code from compressed text descriptions while maintaining a high level of functional accuracy. 

The following is a summary of key lessons learned from the research presented in this paper:

\begin{itemize}
\item \textit{Evaluation metrics provide a sound basis for comparisons} Our SRE and ERE metrics provide a sound and standardized means of assessing the effectiveness of LLM compression techniques, considering both the semantic aspects and the precise textual content of LLM outputs on compression directives. These metrics support future research on LLM compression by fostering the development of more data-efficient models, and contributing to improved interpretability of LLM outputs.
%This has profound implications for the development and deployment of LLMs in various applications, as it can lead to increased efficiency and scalability.

\item \textit{Evaluations are limitation by resource constraints} 
The results presented in this paper are inherently limited by a small number of data samples, as well as limited resources. A principal concern with LLMs is the immense amount of resources required to systematically evaluate bias in outputs at large-scale. 

\item \textit{Reproducibility challenges across releases} The LLM models are updated and modified on an unknown basis, which may affect the reproducibility of our results over time. In a similar vein, different users may observe different model outputs for identical input prompts, for reasons that are not entirely clear, which may limit reproducibility in some cases.
\end{itemize}

%The insights gained from this study have broad applications in both academia and research, offering a solid foundation for further exploration in this field.

% In conclusion, the findings of this paper contribute significantly to the understanding of LLM compression and its potential impact on the efficiency and scalability of these models. 

The following is a non-comprehensive summary of questions that may be of interest for future work in the area of approximate compression with LLMs:
%broad areas and suggested directions that we would like to address in future work.

\begin{enumerate}
    \item {For how long of an inference period can recalled compressed text persist? } 

    % The persistence of compression across an inference period depends on the model architecture, compression techniques employed, and the complexity of the information being compressed. Longer inference periods may result in information degradation, but novel approaches to LLM-based compression could improve the stability and persistence of compressed data, opening up new possibilities for efficient long-term information storage and retrieval.
    % \\

    \item {What meta-representations (e.g., unicode, emoji, math symbols, etc.) are LLMs biased towards, if any? Similarly, what is the derived relative information density of each meta-representation when used by an LLM?} 

    %  and how robust are they between inference periods? 

    % LLM biases towards certain meta-representations may stem from the nature of their training data. Although biases may exist, it is crucial to evaluate the robustness of these meta-representations between inference periods and to quantify their relative information density. Identifying these biases and characterizing their impact on LLM performance can inform the development of more balanced and effective compression techniques.
    % \\
    
    \item {What is the minimal prompt that achieves a compression rate of X for a specific type of data?} 

    % Determining the minimal prompt that achieves a specific compression rate requires an understanding of the interplay between the LLM's architecture, the compression algorithm, and the type of data being compressed. Identifying this minimal prompt can improve the efficiency of information transmission and storage, enabling more effective communication and collaboration between LLMs.
    % \\

    \item {How can cooperating/competing LLMs collectively leverage or refine a contextual compression policy?}

    % Cooperating or adversarial LLMs can leverage and improve contextual compression policies by exchanging compressed data, learning from each other's experiences, and identifying weaknesses in the compression process. By doing so, they can enhance the effectiveness and robustness of compression techniques, enabling more efficient information storage and retrieval in a variety of contexts.
    % \\
    
    \item {Can compressed data be embedded into an existing LLM system in an adversarial way to produce undesired behavior?}

    % Compressed text has the potential to be embedded into existing systems as part of adversarial attacks. This highlights the importance of developing robust security measures and data handling protocols to protect against such attacks and to ensure the integrity and privacy of sensitive information.
    % \\

    \item {Can approximate compression policies be effectively shared or communicated across models?}

    % Compression can be effectively shared across models, enabling them to benefit from each other's experiences and knowledge. This collaborative approach can lead to more efficient and effective compression techniques, as well as improved generalization and adaptability to different tasks and domains.
    % \\

    \item {Can different contextual categories (e.g., code, prose) be derived to taxonomize LLM compression capabilities?}

    % Contextual categories can be derived to classify LLM compression performance and ability, providing insight into how different models excel in specific tasks or domains. Such categorization can inform model selection, optimization, and deployment strategies, leading to more effective and targeted applications of LLM-based compression.
    % \\

    \item {How does the information integrity degrade when recursively compressed by multiple LLM instances?}
    %and what conclusions can be drawn because of it?}

    % The informational integrity of a compression may degrade when compoundingly compressed, potentially leading to information loss or inaccuracies. By analyzing the behavior of compound compression, researchers can identify the limitations and thresholds of LLM-based compression techniques, informing the development of more robust and reliable methods for preserving informational integrity.

\end{enumerate}

As LLMs continue to evolve and serve a wide range of real-world use cases, we believe the insights and metrics presented in this study will serve researchers and practitioners alike, helping to drive the development of more effective and efficient LLMs.

%\section*{Acknowledgment}
% Guys, I'm nervous about saying this since I think it'll undermine our credibility..
% In the spirit of demonstrating the concept of prompt patterns, \textbf{ChatGPT was used extensively in the writing of this paper} with the following prompt: "From now on, whenever I provide a list of bullets, you will expand those bullets into a full paragraph. Each paragraph will have at most 4 bullets. If I provide more than four bullets, you will automatically expand the bullets into multiple paragraphs so that no paragraph covers more than four bullets. You will group related ideas together into a paragraph and each paragraph will have a topic sentence and supporting statements. Every paragraph will have at least three sentences and a maximum of five sentences. When writing, use an academic research paper in computer science tone. Avoid hyperbolic language and claiming benefits and instead state facts."

\bibliographystyle{IEEEtran}
\bibliography{IEEEabrv,bibfile}

\end{document}